\documentclass[letterpaper]{article} 
\usepackage{aaai2026}  
\usepackage{times}  
\usepackage{helvet}  
\usepackage{courier}  
\usepackage[hyphens]{url}  
\usepackage{graphicx} 
\usepackage{natbib}  
\usepackage{caption} 
\usepackage{amsmath}
\usepackage{amsthm}
\usepackage{amssymb}
\usepackage{subfigure}
\usepackage{multirow}
\usepackage{tabularx}
\usepackage[capitalize]{cleveref}
\usepackage{mathrsfs}
\usepackage{cases}
\usepackage{enumerate}
\usepackage{algpseudocode}
\usepackage{booktabs}
\usepackage[switch]{lineno}
\usepackage{multirow}
\usepackage{tabularx}
\usepackage[capitalize]{cleveref}
\usepackage{mathrsfs}
\usepackage[linesnumbered,ruled,vlined]{algorithm2e}

\newtheorem{lemma}{Lemma}
\newtheorem{definition}{Definition}

\newtheorem{theorem}{Theorem}

\newtheorem{assumption}{Assumption}

\usepackage{enumitem}
\setlist[enumerate,1]{label=(\arabic*).,font=\textup,
leftmargin=8mm}

\usepackage{newfloat}
\usepackage{listings}
\DeclareCaptionStyle{ruled}{labelfont=normalfont,labelsep=colon,strut=off} 
 \nocopyright

\title{FedCure: Mitigating Participation Bias in Semi-Asynchronous Federated Learning with Non-IID Data}
\author{
    Yue Chen\textsuperscript{\rm 1},
    Jianfeng Lu\textsuperscript{\rm 2,\rm 3}\thanks{Corresponding Authors are Jianfeng Lu and Guanghui Wen.},
    Shuqing Cao\textsuperscript{\rm 2},
    Wei Wang\textsuperscript{\rm 1},
    Gang Li\textsuperscript{\rm 4},
    Guanghui Wen\textsuperscript{\rm 5}\footnotemark[1]
}
\affiliations{
    \textsuperscript{\rm 1}School of Computer Science and Technology, Wuhan University of Science and Technology, China\\
    \textsuperscript{\rm 2}2.Hubei Province Key Laboratory of Intelligent Information Processing and Real-time Industrial System, Wuhan University of Science and Technology, China\\
    \textsuperscript{\rm 3}Key Laboratory of Social Computing and Cognitive Intelligence (Dalian University of Technology), Ministry of Education, China\\
    \textsuperscript{\rm 4}College of Computer Science, Inner Mongolia University, China\\
    \textsuperscript{\rm 5}School of Automation, Southeast University, China\\
    \{chenyue, lujianfeng, shuqincao,wangwei8\}@wust.edu.cn, gli@imu.edu.cn, ghwen@seu.edu.cn
}

\usepackage{bibentry}

\begin{document}

\maketitle

\begin{abstract}
While semi-asynchronous federated learning (SAFL) combines the efficiency of synchronous training with the flexibility of asynchronous updates, it inherently suffers from participation bias, which is further exacerbated by non-IID data distributions. More importantly, hierarchical architecture shifts participation from individual clients to client groups, thereby further intensifying this issue. Despite notable advancements in SAFL research, most existing works still focus on conventional cloud-end architectures while largely overlooking the critical impact of non-IID data on scheduling across the cloud–edge–client hierarchy. To tackle these challenges, we propose FedCure, a innovative semi-asynchronous \underline{Fed}erated learning framework that leverages \underline{C}oalition constr\underline{u}ction and pa\underline{r}ticipation-aware sch\underline{e}duling to mitigate participation bias with non-IID data. Specifically, FedCure operates through three key rules: (1) a preference rule that optimizes coalition formation by maximizing collective benefits and establishing theoretically stable partitions to reduce non-IID-induced performance degradation; (2) a scheduling rule that integrates the virtual queue technique with Bayesian-estimated coalition dynamics, mitigating efficiency loss while ensuring mean rate stability; and (3) a resource allocation rule that enhances computational efficiency by optimizing client CPU frequencies based on estimated coalition dynamics while satisfying delay requirements. Comprehensive experiments on four real-world datasets demonstrate that FedCure improves accuracy by up to 5.1x compared with four state-of-the-art baselines, while significantly enhancing efficiency with the lowest coefficient of variation 0.0223 for per-round latency and maintaining long-term balance across diverse scenarios.
\end{abstract}

\section{Introduction}\label{intro}
According to Statista, a leading market data provider, the number of IoT devices worldwide is projected to approach 30 billion by 2030 \cite{Statistica}. The rapid proliferation of devices is accelerating the development of edge intelligence services such as smart healthcare \cite{smart_health}, Internet of Vehicles \cite{IoV}, and smart manufacturing \cite{smart_manufacturing}, but faces data privacy leakage and communication overload. Federated Learning (FL) \cite{FL1}, a distributed machine learning paradigm, addresses privacy concerns by enabling local training and exchanging only model parameters. To further reduce communication latency and improve system efficiency, the semi-asynchronous FL (SAFL) framework with hierarchical architectures \cite{HFL_compute} has gained increasing attention. By combining synchronous aggregation at the client-edge layer with asynchronous updates at the edge-cloud layer \cite{SAFL_step}, SAFL alleviates the traditional cloud-end communication bottleneck and significantly improves training efficiency, demonstrating strong potential across diverse distributed intelligent applications.

However, the semi-asynchronous nature of SAFL, combined with heterogeneous client responsiveness in real-world deployments, gives rise to a critical issue of participation bias \cite{monoply}. Under asynchronous training, clients’ participation frequency heavily depends on its responsiveness. Models from faster-responding clients are frequently aggregated, while the slower clients are marginalized over time. This skewed participation leads to model updates being dominated by a few clients, thereby degrading model generalization. This issue not only compromises system performance but also contradicts the core of AI development. The updated “AI Principles” of OECD  in 2024 \cite{OECD} and the 2025 Paris AI Action Summit \cite{AI_action_summit} all emphasize “Inclusive Growth”.

Although some studies have attempted to tackle participation bias by adding restrictions based on device heterogeneity \cite{SAFL_1}, the asynchronous nature of SAFL causes highly dynamic participation, making static modeling insufficient to capture clients' real-time capabilities. Moreover, participation bias often intertwines with additional challenges, such as non-IID data \cite{SAFL_2}, which warrant further investigation. Current researches mainly focus on the cloud-end structure, while limited exploration of edge-end and edge-cloud collaboration, hindering the practical deployment of SAFL \cite{HFL_drawback}.

The opposite of participation bias is balanced scheduling, which can improve model comprehensiveness and generalization \cite{SM3,faironly}. However, achieving this in SAFL faces three key challenges: 1) \textit{Non-IID data exacerbates participation bias.} Non-IID data further amplifies certain data features while neglecting other critical information during participation biased training \cite{noniid1}. 2) \textit{Hierarchical architecture complicates scheduling.} Cloud Server (CS) aggregates edge models from groups of clients associated with the same Edge Server (ES), shifting the scheduling focus from individual clients to coalitions \cite{noniid_set}. 3) \textit{Efficiency degradation under fully balanced scheduling.} Balanced scheduling inevitably reduces efficiency due to the participation of high-latency clients, which necessitates a trade-off \cite{trade_off}.

To address these challenges, we propose FedCure, a novel SAFL framework that mitigates participation bias under non-IID data while ensuring model performance and training efficiency. Specifically, we first formalize the challenges as three constraints: the edge association constraint affects data distribution by restricting the client-ES association (Challenge 1), the scheduling constraint imposes long-term participation requirements on coalitions (Challenges 2 \& 3), the resource constraint promotes feasible client contribution for efficiency (Challenge 3). FedCure addresses these constraints via three key rules: (i) We construct a coalition formation game to model the client-ES association and design \textit{a preference rule} to minimize the Jensen-Shannon Divergence, thereby reducing non-IID level and satisfying the edge association constraint. (ii) To balance participation and ensure efficiency, we design \textit{a scheduling rule} by modeling the participation frequency of coalitions and combining coalition dynamics (i.e., execution delay). (iii) To further improve training efficiency, we propose \textit{a resource allocation rule} that sets the optimal CPU frequency for clients based on coalition dynamics, ensuring compliance with delay requirements. The specific contributions are as follows:
\begin{itemize}
    \item Methodologically, prior to training, we propose a preference rule for coalition formation that prioritizes collective benefit, avoiding inefficiencies of selfish and Pareto preferences and mitigating model degradation under non-IID data. During training, to counter efficiency loss from fully balanced participation, we employ Bayesian estimation to capture coalition dynamics and integrate a virtual queue to guide scheduling for more efficient training. Building on the above, we introduce a resource allocation rule to dynamically set optimal CPU frequencies within feasible bounds, further improving efficiency.
    
    \item Theoretically, we prove that FedCure guarantees the existence of a stable coalition partition, as the coalition formation game based on the proposed preference rule is proven to be an exact potential game. Moreover, we show that FedCure guarantees the mean rate stability of the virtual queue, thereby achieving long-term balance under SAFL. We also derive an upper bound on the error induced by the trade-off between efficiency and balance.
    
    \item Experimentally, we verify the effectiveness of FedCure on four real-world datasets, comparing it with two clustering and two scheduling methods, and observe a remarkable 5.1x improvement in accuracy. FedCure aslo attains the lowest coefficient of variation for per-round training latency, with a value of 0.0223. Furthermore, we confirm that FedCure maintains long-term balance under varying weight settings.
\end{itemize}

\section{Related Work}
\paragraph{Non-IID Data in SAFL.} In FL, data heterogeneity across clients leads to non-IID distributions, causing local models to drift from the global optimum and ultimately degrading model performance \cite{NonIID}. \citeauthor{kmeans} \shortcite{kmeans} and \citeauthor{meanshift} \shortcite{meanshift} utilized the K-Means and Mean-Shift algorithms, respectively, to cluster clients and selectively sample participants from different clusters for training. Distinguishing from the clustering approach, \citeauthor{share_data} \shortcite{share_data} mitigated the effects of non-IID by sharing a portion of the coalition master's data. \citeauthor{HFL_drawback} \shortcite{HFL_drawback} explored the non-IID challenge in SAFL and proposed a heterogeneity-aware client-edge association mechanism, which enable fast and accurate cloud model learning by measuring response latency and collecting clients' data distributions. However, the IID assumption of global data distribution presents limitations in real-world scenarios.

While the client selection strategy after clustering mitigates the non-IID problem to some extent, it still excludes clients with rare or unique data from  training. In contrast, our work focuses on client coalitions, which naturally aligning with the hierarchical structure of SAFL. Moreover, our work continuously guides the system toward a more favorable data distribution without relying on additional assumptions or introducing the uncertainty of random selection, ultimately achieving a stable and efficient coalition structure.

\paragraph{Scheduling Strategy in SAFL.} Given the limited resources and heterogeneous capability, effective scheduling algorithms have been increasingly studied to enhance model performance and training efficiency. In synchronous FL, \citeauthor{SM2} \shortcite{SM2} prioritized efficient clients to accelerate convergence, while \citeauthor{SM5} \shortcite{SM5} maximized expected cumulative effective participation to improve convergence speed. In asynchronous FL, \citeauthor{SM3} \shortcite{SM3} proposed a dynamic client selection approach to reduce the overall training duration. With the advent of SAFL, related works emerged, such as \citeauthor{SAFL_1} \shortcite{SAFL_1} proposed a clustered SAFL that groups devices with similar training conditions to mitigate the impact of stragglers. \citeauthor{SAFL_2} \shortcite{SAFL_2} proposed Libra, which addresses unfair client scheduling by alleviating the limitations of fast-training devices due to excessive training and upload demands.

Favoring specific clients in scheduling may exhaust their resources, leaving out others that possess important data. Moreover, the impact of non-IID data underlying participation bias also deserves serious attention. In our study, scheduling is based on coalitions with data that exhibits a high level of IID. The scheduling rule jointly considers participation balance and training efficiency, with further efficiency gains achieved via resource optimization.

\section{System Model}
\subsection{Workflow of SAFL}
A typical SAFL system consists of a set $ {\cal N} = \left\{ {1, \cdots, N} \right\} $ of clients, a set ${\cal M} = \left\{ {1, \cdots, M} \right\}$ of ESs, and a CS. The dataset of the client $n$ is denoted as $\mathcal{D}_n =\left \{ \mathcal{X}_n ,\mathcal{Y}_n  \right \} $, where $\mathcal{X}_n =\left \{x_{n,1}, \cdots , x_{n,|D_n|} \right \} $ is the training dataset, $\mathcal{Y}_n =\left \{y_{n,1}, \cdots , y_{n,|D_n|} \right \}$ is the corresponding label set. We define ${\cal G}_m$ as the set of clients that are associated with ES $m$, which form a coalition. Since each client can only be associated with one ES, the coalitions are disjoint, i.e., ${{\cal G}_m} \cap {{\cal G}_{m'}} = \emptyset$ for $m \ne m'$. The training of SAFL occurs in the following two layers:

1) {\bf Client-Edge Layer:} Each client in coalition ${\cal G}_m$ receives the interim model from ES $m$ to train a local model using its dataset \cite{SM}, i.e.,
${F} _{n}\left ( \omega _{n}  \right )  =\frac{1}{\left | \mathcal{D}_{n}  \right | } \sum_{i\in \mathcal{D}_{n}} f_{i}\left ( \omega _{n}\right )$.
After every $\tau_c$ rounds of local updates, each client transmits its updated model ${\omega}_{n,m}^{t_e}$ to ES $m$ during the $t_e$-th edge training round. Upon receiving the local models from all clients in ${\cal G}_m$, the ES $m$ proceeds with the edge aggregation process. The updated edge model is defined as
\begin{equation}
   \omega_m^{t_e+1} =\frac{ {\textstyle \sum_{n\in\mathcal{G}_{m} }}\left | \mathcal{D}_{n}  \right | \omega _{n,m}^{t_e} }{ {\textstyle  \left | \mathcal{D}_{m}  \right |} },
    \label{edge_agg}
\end{equation}
where $\left |{\cal D}_m\right | = \sum_{n \in{\cal G}_m} \left |{\cal D}_n\right | $ is the sample data size of all clients in ${\cal G}_m$. After $\tau_e$ rounds of edge updates, the edge model will be forwarded to the CS for global aggregation.

2) {\bf Edge-Cloud Layer:} The CS performs global aggregation as soon as it receives a model from any ES without waiting. However, asynchronous aggregation introduces the challenge of staleness, which may degrade the convergence rate and in turn increase the training time. For coalition ${\cal G}_m$, its staleness is the number of experienced epochs since its last global update, denoted as $\varphi$. For example, when the CS receives a stale model at epoch $t$ from ES $m$, the updated global model will be determined by $\varphi$ \cite{aggregation}:
\begin{equation}
   \omega^{t} =\left ( 1-\xi_{\varphi}\right ) \omega^{t-1}+\xi_{\varphi}\omega_m,
    \label{edge_agg}
\end{equation}
where $\xi_{\varphi }= \ell \cdot  \Bbbk^{\varphi}$ is the weight of $\omega_m$ with staleness $\varphi$, $\ell \in (0,1)$ is initial weight, and $\Bbbk \in (0,1)$ is penalty coefficient. Intuitively, the model with smaller staleness has a larger $\xi_{\varphi }$ and contributes more to model aggregation.

For a global round $t$, a coalition will be selected before the CS sends the latest model. The coalitions that have not uploaded their trained models are considered unavailable in our work. In particular, we define $\theta_m(t)=1$ if coalition is available, and $\theta_m(t)=0$, otherwise. Hence, the optional set of coalitions in global round $t$ is $\Theta(t) = \left [ \theta_m(t)\right ]_{1\times M}$.

\subsection{Three Constraints in SAFL}
\subsubsection{Edge Association Constraint (EAC)}
In the client-edge layer, association relationships affect the IID degree of data. A lower Jensen-Shannon Divergence (JSD) \cite{JSD} indicates greater similarity between two data distributions, which means data is more evenly distributed.
\begin{definition}
The average JSD value across the coalitions is calculated as follows:
\begin{equation}
      \mathcal{\overline{JS}} = \frac{1}{\binom{M}{2} }\sum_{i=1}^{M-1} \sum_{j=i+1}^{M} \mathcal{JS}(Q_i, Q_j), \label{JS散度计算_Avg}
\end{equation}	
where $\mathcal{JS}(Q_i, Q_j) = \frac{1}{2}\left\{ \mathcal{KL}(Q_i, \mathbb{M}_{ij}) + \mathcal{KL}(Q_j, \mathbb{M}_{ij})\right\}$, $Q_i$ is the probability distribution of data within coalition ${\cal G}_i$, $\mathcal{KL}(\cdot)$ is Kullback-Leibler Divergence \cite{KLD}, and $\mathbb{M}_{ij}=\frac{Q_i+Q_j}{2}$ is the mean distribution. 
\end{definition}

To achieve a lower level of non-IID distribution, the formed coalitions need to satisfy the following constraint, ensuring that no client's switching can reduce $\mathcal{\overline{JS}}$: 
\begin{equation}
	\mathcal{\overline{JS}} < \mathcal{\overline{JS}}_{\mathcal{G}_b \to\mathcal{G}_a}^n, \forall n \in N,
\end{equation}
where $\mathcal{\overline{JS}}_{\mathcal{G}_b \to\mathcal{G}_a}^n$ is the average JSD value after client $n$ leaves original coalition $\mathcal{G}_b$ to join the new coalition $\mathcal{G}_a$.

\subsubsection{Scheduling Constraint (SC)} 
To avoid local optima in SAFL, we introduce a long-term balance constraint on the participation requirement \cite{longfair}, thereby achieving relative balance among coalitions and guaranteeing a minimum number of participation rounds for each coalition, 
\begin{equation}\label{longterm}
   \underset{\tau_g \rightarrow \infty  }{\lim } \frac{1}{\tau_g} \sum_{t=0}^{\tau_g-1} \mathbb{E} [\chi_m(t)]\ge \delta _m,\delta _m \in (0,1], \forall m\in\cal M,
\end{equation}
where $\delta_m$ is the minimum fraction that coalition $m$ should participate in model training, $\mathcal{X}(t)=[ \chi _1(t),\cdots ,\chi_M(t) ]^T $, and $\chi_m(t)$ is a binary variable indicating whether coalition $m$ is scheduled in round $t$. Specifically,  if CS schedules coalition $m$, then $\chi_m(t)=1$, $\pi(t)=m$, and $\chi_m(t)=0$ otherwise. $\pi(t)$ is the index of the scheduled coalition in round $t$ and $\Pi$ is the set of $\pi(t)$ of all rounds. In addition, $\sum_{m=1}^{M} \chi_m(t)=1,t\ge 1$, which means only one coalition can be scheduled in each round. The CS schedules $M$ coalitions in the 0-th round for initialization. We set the boundary of the expected scheduling probability as $\delta_m=\frac{\kappa \left |{\cal D}_m \right |}{\left | \cal D \right |}$, $\kappa \in [0,1]$, where $\left |\cal D\right |$ is the data size of all coalitions.

\subsubsection{Resource Constraint (RC)} Clients in the scheduled coalition ${\cal G}_{\pi(t)}$ perform synchronized training, facing a trade-off between system efficiency and resource consumption.
\begin{equation}
  \mathcal{Z}=\mathbb{E} \left \{ \alpha (1-\frac{t_n}{\mathcal{T}_m(t)} ) - C_n(f_n) \right \},
\end{equation}
where ${\cal T}_m(t)$ is the training latency, $ t_n = c_n / f_n$ is the computation time, $c_n$ is the computation load, $f_n \in (0,f_n^{max}]$ is the computation resource allocated by client $n$, $C_n(f_n) = \gamma (f_n )^ \varsigma$ is the energy consumption of client $n$ in model training \cite{comp_cost}, and $\alpha$, $\gamma$, $\varsigma$ are weight coefficients.

\subsection{Problem Formulation}
High-quality data improve model performance, whereas prolonged delays and high energy consumption bring negative impacts. Based on EAC, SC and RC, we model the problem to optimize efficiency and scheduling in SAFL:
\begin{subequations}
\begin{align}
{\mathbb{P} }:& \quad \underset{{\left\{ {{\cal G}} \right\}_{1}^{M}}, \pi(t)}{\min } \sum_{t=0}^{\tau_g}\frac{\mathcal{T} _{\pi(t)}(t)}{\mathcal{I}} \quad \& \quad \underset{f}{\max} \enspace \mathcal{Z}, \label{P1}\\
& \quad \text{s.t.}\quad (4),\enspace(5) \enspace \text{and} \enspace (6),\label{eq:constraint2}\\
& \quad \quad \quad \pi(t) \in {\bf m}^T\Theta (t),\label{eq:constraint2}\\
& \quad \quad \quad  f_n \in (0,f_n^{max}],\label{eq:constraint4}\\
& \quad \quad \quad m = \pi(t), \enspace n\in {\cal G}_{\pi(t)},  \label{eq:constraint3}
\end{align}
\end{subequations}
where ${\cal G}_{\pi(t)}$ is the scheduled coalition at round $t$, ${\bf m}=[1, \cdots,M]^T$, and ${\cal I}$ is the average max training latency. 

This problem shows a profound interdependence between constraints, presenting two-fold challenges: (i) SC and RC must be solved jointly under the satisfied EAC, with RC depending on the coalition scheduled by SC, forming a tightly coupled process; (ii) the time-coupled SC further increases the complexity. 

\begin{figure}[t]
\centering
 \includegraphics[scale=0.286]{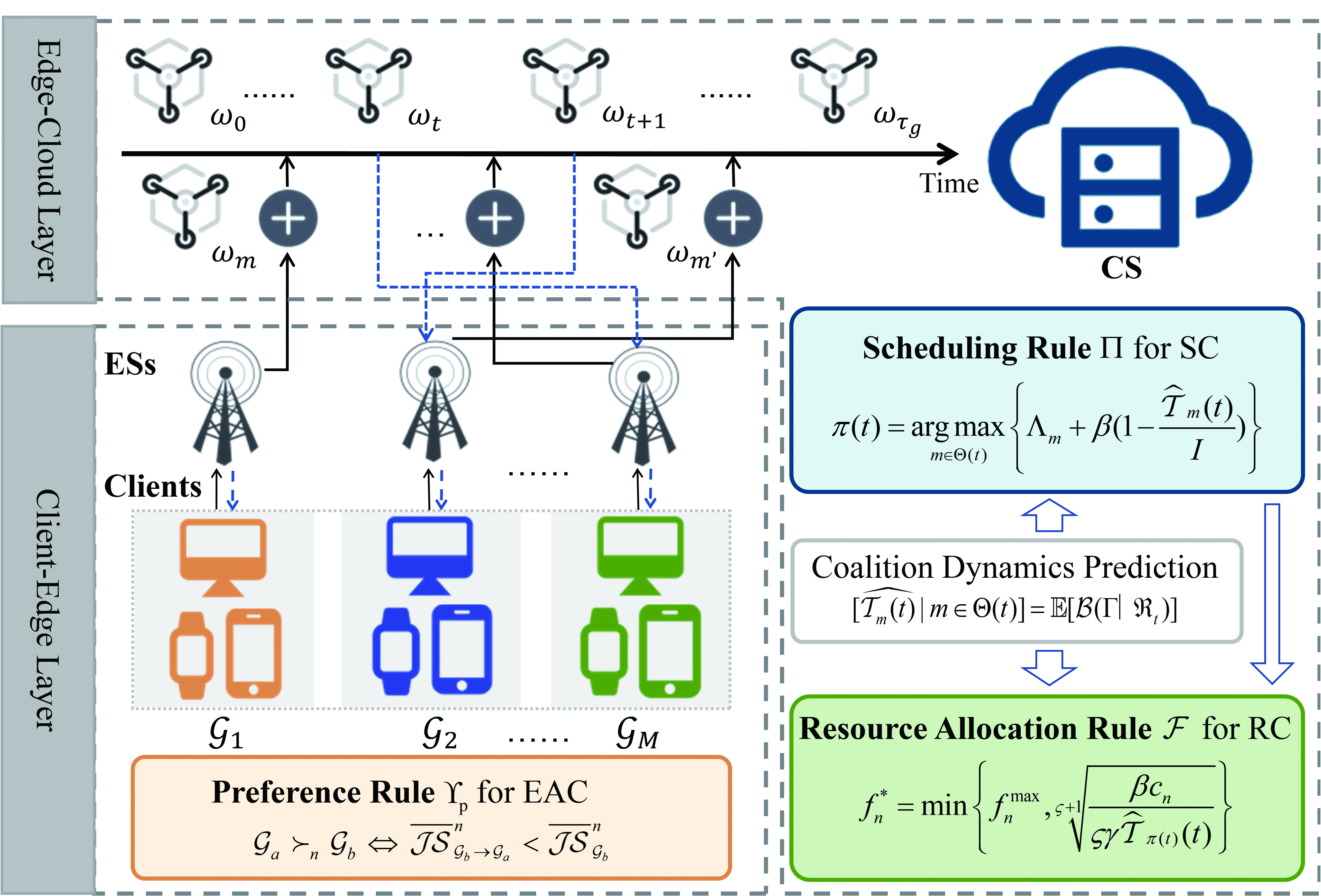}
\caption{An overview of FedCure.}
\label{system_model}
\end{figure}

\section{Design of FedCure}
In this section, we address problem $\mathbb{P}$ by designing FedCure with three strategic rules.
\begin{definition}
FedCure is defined as a 3-tuple $(\Upsilon_p, \Pi, \cal{F})$:
\begin{itemize}
	\item $\Upsilon_p \Rightarrow \cal O$: a preference rule that guides a coalition partition $\cal O$ formulating by determining client preferences over ESs, aiming to mitigate non-IID issue.
	\item $\Pi = \bigcup_{t=1}^{\tau_g} \left \{ \pi(t) |\pi(t) \in {\bf m}^T\Theta (t) \right \} $: a scheduling rule $\Pi$ governing coalition participation ${\cal G}_{\pi(t)} \in {\cal O}$ per global round $t$, considering the trade-off between balanced scheduling and training efficiency.
	\item ${\cal F}=\bigcup_{t=1}^{\tau_g } \left \{ f_n| n \in \pi(t) \right \}$: a resource allocation rule $\cal{F}$ sets the CPU frequency $f_n$ for clients in scheduled coalition ${\cal G}_{\pi(t)}$ to optimize computational capacity.
\end{itemize}
\end{definition}
\paragraph{Remark:} As illustrated in Figure~\ref{system_model}, the preference rule $\Upsilon_p$ guides clients to associate with ESs, forming stable coalition partitions that satisfy the EAC by reducing the average JSD value. During asynchronous training, the CS applies the scheduling rule to select participating coalitions, ensuring long-term balance as required by the SC, while maintaining training efficiency. Furthermore, clients in selected coalitions train at CPU frequencies determined by the resource allocation rule, further improving efficiency in compliance with the RC.

\subsection{Optimal Design of Preference Rule}
To meet EAC, we construct a coalition formation game \cite{CFG} where clients dynamically associate with ESs under the preference rule $\Upsilon_p$ that minimizes $\overline{JS}$ across ESs, forming a stable partition with high IID degree.

\begin{definition}
    A coalition formation game $\mathbb C$ is represented as a 4-tuple ($\cal N$, $\cal O$, $\Upsilon_p$, $\Upsilon_s$), i.e.,
    \begin{itemize}
        \item $\cal N$: A set ${\cal N} = \left \{ 1, \cdots, N  \right \} $ of players.

	   \item $\cal O$: A coalition partition ${\cal O} = \left\{ {{\cal G}} \right\}_{1}^{M}$, where ${{\cal G}_m} \subseteq {\cal O}$, $ \cup _{m = 1}^{M}{{\cal G}_m} = {\cal N}$.

        \item $\Upsilon_p$: A preference relation  ${\succ_n}$ is a complete, reflexive, and transitive binary relation over the set of all coalitions that client $n$ may join in, i.e., 
      \begin{equation}
	\mathcal{G}_a\succ _n \mathcal{G}_b\Leftrightarrow  \mathcal{\overline{JS}}_{\mathcal{G}_b \to\mathcal{G}_a}^n<  \mathcal{\overline{JS}}_{\mathcal{G}_b}^n,
	\end{equation}
where ${{\cal G}_a}{ \succ _n}{{\cal G}_b}$ indicates that client $n$ strictly prefers to join coalition ${{\cal G}_a}$ over coalition ${{\cal G}_b}$, $ \mathcal{\overline{JS}}_{\mathcal{G}_b}^n$ and $\mathcal{\overline{JS}}_{\mathcal{G}_b \to\mathcal{G}_a}^n$ represent the $ \mathcal{\overline{JS}}$ values before and after client $n$ leaves the coalition ${\cal G}_b$, respectively.

        \item $\Upsilon_s$: A switching rule states that a client $n \in {{\cal G}_a}$ decides to leave ${{\cal G}_a}$ and join ${{\cal G}_{b}}$ ($b \ne a$) if and only if ${{\cal G}_{b}} \cup \left\{ n \right\}{\underline  \succ  _n}{{\cal G}_{a}}$ holds, and hence the new coalition partition is updated to
         \begin{equation}
	\widetilde {\cal O} \!\to\! \left\{ {\left( {{\cal O} \backslash\! \left\{ {{{\cal G}_a},{{\cal G}_b}} \right\}} \right) \!\cup\! \left( {{{\cal G}_a} \!\backslash\! \left\{ n \right\}} \right) \!\cup\! \left( {{{\cal G}_b} \!\cup\! \left\{ n \right\}} \right)} \right\}.
	\end{equation}       
    \end{itemize}
\end{definition}
There are two widely used preference rules: the ``Self-Centered Rule" \cite{SHFL} prioritizes individual choices but risks coalition cohesion, while the ``Pareto Rule" \cite{pareto} safeguards against harm but may constrain growth. In contrast, we propose a coalition-friendly preference rule $\Upsilon_p$ that emphasizes the welfare of coalition partition. Clients seek associations by calculating the reduction of $\mathcal{\overline{JS}}$ before and after the switch. $\Upsilon_p$ is assessed from a coalition perspective, similar to partial collaboration. Consequently, examining its stability becomes paramount.

\begin{definition}
An exact potential game exists when a potential function $\phi$ maintains a constant difference from the utility function, regardless of client association changes, i.e.,
    \begin{equation}
        \phi( \widetilde{a_n},\!a_{-n})- \phi(a_n,\!a_{-n}) \!=\! {\cal U}_n(\widetilde{a_n},\!a_{-n}) \!-\! {\cal U}_n(a_n,\!a_{-n}),
    \end{equation}
where $a_n$ denotes the coalition that client $n$ joined in, and $a_{-n}$ represents the associations of all other clients except $n$.
\end{definition}
\begin{algorithm}[tb]
    \caption{Data Distribution Adjustment}
    \label{alg:algorithm1}
  \KwIn{Client set ${\cal N}$, ES set ${\cal M}$, current partition ${{\cal O}_{cr}}$, and maximum iteration round $L$}
  \KwOut{Final partition ${{\cal O}^*}$}
  Initialize ${{\cal O}^*} = \emptyset$, $l = 0$\;
  \Repeat{coalition partition converges or $l = L$}{
	$n = \text{Random}\left\{1, \cdots, N\right\}$, $n\in {\cal{G}}_m$\;
	\ForEach{${{\cal G}_{m'}} \in {{\cal O}_{cr}}, m \ne m'$}
	{
		Calculate $ \mathcal{\overline{JS}}_{\mathcal{G}_m \to \mathcal{G}_{{m}'}}^n$ according to Eq. (\ref{JS散度计算_Avg})\;
	}
	${m}' =\underset{m}{\min} \left \{  \mathcal{\overline{JS}}_{\mathcal{G}_m \to\mathcal{G}_1}^n, \cdots, \mathcal{\overline{JS}}_{\mathcal{G}_m \to\mathcal{G}_M}^n\right \} $\;
	\If{${m}' != m$}
	{
		${{\cal G}_m} = {{\cal G}_m}\backslash \left\{ n \right\}$, ${{\cal G}_{m'}} = {{\cal G}_{m'}} \cup \left\{ n \right\}$\;
		${{\cal O}_{cr}} = \left( {{{\cal O}_{cr}}\backslash \left\{ {{{\cal G}_m},{{\cal G}_{m'}}} \right\}} \right) \cup \left( {{{\cal G}_m}\backslash \left\{ n \right\}} \right) \cup \left( {{{\cal G}_{m'}} \cup \left\{ n \right\}} \right)$\;
	}
	 $l = l+1$, ${\cal O}^* = {\cal O}_{cr}$\;
}
\end{algorithm}
\begin{theorem}
    The coalition formation game $\mathbb{C}$, guided by the preference rule $\Upsilon_p$, is an exact potential game that can form a stable partition to satisfy EAC.
    \label{EPG}
\end{theorem}

Algorithm~\ref{alg:algorithm1} illustrates the process of constructing a stable partition. Specifically, $\Upsilon_p$ evaluates the highest preference of the selected client $n$ ($n \in {\cal G}_m$) among coalitions (lines 3-6). We assume client $n$ leaves its current coalition and calculates $\mathcal{\overline{JS}}$ for each potential scenario of joining a different coalition using Eq.~(\ref{JS散度计算_Avg}) (line 5), allowing prioritization based on the resulting values (line 6). Lines 7-9 show the switch operation of clients between coalitions when $ \mathcal{\overline{JS}}$ increased, otherwise, the client remains in its current coalition. This iterative process continues until a stable partition ${{\cal O}^*} = \left\{ {{\cal G}^*} \right\}_1^M$ is formed, where no exchange can further reduce $ \mathcal{\overline{JS}}$ or the maximum number of iteration rounds is reached. Calculating $ \mathcal{\overline{JS}}$ requires $\frac{\left ( M-1 \right ) M}{2}$ steps, yielding a total time complexity ${\cal O}\left ( M^2\right )$. It's crucial to note that the computation is distributed across all ESs in reality, so each ES only computes $(M-m)$ times for $\sum_{j=m+1}^{M} \mathcal{JS}(Q_m, Q_j)$, reducing the time complexity to ${\cal O}\left ( M\right ) $. This workload is a negligible load for high-performance ESs.

\subsection{Near-Optimal Scheduling Rule Design and Optimal Resource Allocation Rule Design}
\subsubsection{Coalition Dynamic Estimation} The solution of $\mathbb P$ depends on current delay $\mathcal{T}_m(t)$, requiring dynamic prediction. 
While the law of large numbers states event frequency approximates probability as the number of occurrences increases \cite{large_num}, this fails with limited rounds and scarce data. To address this, we employ Bayesian estimation \cite{bayesian} to estimate the latency vector $\Gamma  = [ \mathcal{T}_m(t) | m \in {\bf m}^T\Theta (t)] $ as follows:
\begin{numcases}{}
	 \mathcal{H} (\mathcal{R}_t \mid \Gamma ) = \prod_{m\in \Theta(t) }\mathcal{P}(r_{t,m}\mid \Gamma ),\\
	 \mathcal{B} (\Gamma \mid \mathcal{R}_t)= \frac{\mathcal{P} (\Gamma )\mathcal{H} (\mathcal{R}_t \mid \Gamma )}{\int_{\Gamma}\mathcal{P} (\Gamma )\mathcal{H} (\mathcal{R}_t \mid \Gamma )d\Gamma },\label{estimate}
\end{numcases}
where $\Gamma$ denotes the parameters that need to be estimated, $\mathcal{P} \left ( \Gamma \right )$ is a prior distribution, and $\mathcal{R}_t =$$ \left \{ r_{t,m} | m\in {\bf m}^T\Theta (t) \right \}$ is the latency vector from $\mathcal{P} \left ( \Gamma \right )$. Based on the observed samples, we construct the joint probability distribution function $\mathcal{H} (\mathcal{R}_t \mid \Gamma )$, calculate the marginal probability $\int_{\Gamma}\mathcal{P} (\Gamma )\mathcal{H} (\mathcal{R}_t \mid \Gamma )d\Gamma $,  and use the Bayesian formula to get the posterior distribution for the parameters $\mathcal{B} (\Gamma \mid \mathcal{R}_t)$.

Based on these steps, we obtain the parameter estimation $[\hat{\mathcal{T}_m(t)} | m\in {\bf m}^T\Theta (t) ]=\mathbb{E}[\mathcal{B} (\Gamma \mid \mathcal{R}_t)]$. Since the CS can estimate the successful probability of model upload of each available coalition, we gain a clearer scheduling direction.

\subsubsection{Coalition Scheduling}
As long-term SC poses difficulties for conventional methods, we adopt the virtual queue approach from Lyapunov optimization, defined as:
\begin{equation}
\!\!\!\!\Lambda_m(t)\!\!=\!\!\left\{
\begin{aligned}
& \!-\!\delta_m, &\!\!\!\!\!\text{if}\!\enspace\!t\!=\!-1,\\
& \!\max \left\{ \Lambda _m(t\!-\!1)\!+\!\delta _m \! -\!\chi_m(t), 0\right\}, &\!\!\!\!\!\text{otherwise.}\\
\end{aligned}
\right.
\label{queue}
\end{equation}
The CS is more intuitively referenced for coalition selection based on the estimated execution time $\mathbb{E}[\mathcal{B} (\Gamma \mid R_t)]$ and the virtual queue $\Lambda_m(t)$. Consequently, $\pi (t) $ in $\mathbb P$ can be converted into a solution that satisfies the following scheduling rule $\pi(t) \in \Pi, t>0$:
\begin{equation}
\pi(t)=\underset{m\in {\bf m}^T\Theta (t)}{\arg\max}  \left \{ \Lambda_m (t) + \beta ( 1 - \frac{\hat{\mathcal{T}}_m(t)}{\cal I} )  \right \}, 
\label{selection_rule}
\end{equation}
where $\beta > 0$ is a weight parameter.

If the long-term balance of scheduling constrain in Eq.~(\ref{longterm}) is violated, the virtual queue grows unbounded. To ensure long-term balance, each coalition's queue length must satisfy the following defined mean rate stability \cite{virtual_queue}.
\begin{definition}
	Discrete-time queue $\Lambda(t)$ is mean rate stable if $\underset{t \to \infty}{lim}\frac{{\mathbb E}[\Lambda(t)]}{t}=0$.
	\label{rate_stable}
\end{definition}
\begin{theorem}
FedCure ensures the mean rate stability of the virtual queues, i.e., 
\begin{equation}
	\underset{\tau_g \to \infty}{lim}\frac{{\mathbb E}[\Lambda(\tau_g)]}{\tau_g}=0,
\end{equation}
and satisfies the long-term balance constraint. 
\label{theorem_fair}
\end{theorem}
According to Theorem \ref{theorem_fair}, long-term balance is consistently preserved under the proposed scheduling rule. Notably, this balance holds regardless of the value of the weight parameter $\beta$ in Eq. (\ref{selection_rule}). Moreover, Eq. (\ref{selection_rule}) reflects the trade-off between efficiency and virtual queue length in each rounds, and a near-optimal upper bound is analyzed in the next section.

\subsubsection{Resource Allocation Optimization}
The resource allocation rule $\cal{F}$ optimizes CPU frequency for clients in ${\cal G}_{\pi(t)}$ based on the estimated latency.
\begin{theorem}
There exists an optimal solution $f_{n}^* \in {\cal F}, \forall n\in{\cal G}_{\pi(t)}$ for $\mathbb{P}$, i.e.,
\begin{equation}
    f_n^*\!=\!\min\left\{ f_n^{max} \!,\! \sqrt[\varsigma +1]{\frac{\alpha c_n}{\varsigma \gamma \hat{\mathcal{T}}_{\pi(t)}(t)} }\right\}, n \!\in\! {\cal G}_{\pi(t)}.
    \label{opt_f}
\end{equation}
\end{theorem}

 \begin{algorithm}[tb]
    \caption{Scheduling and Resource Allocation}
    \label{alg:algorithm2}
   \KwIn{Latency vector $\mathcal{R}_t$, Available coalitions $\Theta(t)$}
	\KwOut{Scheduled coalition $\pi(t)$, allocated computation power $\left\{f_n^*\right\}, n \in {\cal G}_{\pi(t)}$}
  Initialize $\Lambda_m(-1)=-\delta_m, \chi_m(-1)=0$\;
  \ForEach {$t<\tau_g$} 
  {
  	\ForEach {$m\in\Theta(t)$}
  	{
  		Estimate $[\hat{\mathcal{T}_m(t)} ]$ and update $\Lambda_m(t)$ with Eq. (\ref{estimate}) and Eq. (\ref{queue}), respectively\; 
  	}
  	\eIf{t=0}
  	{
  		Select all the ESs in $\cal M$ for initialization\;
  	}{
  		$\pi(t)=\underset{m\in\Theta(t)}{\arg\max}  \left \{ \Lambda_m (t) + \beta ( 1 - \frac{\hat{\mathcal{T}}_m(t)}{\cal I} )  \right \}$\;
  	}
  	Calculate $\left\{f_n^*\right\}, n \in {\cal G}_{\pi(t)}$ according to Eq. (\ref{opt_f})\;
  }
\end{algorithm}
Algorithm \ref{alg:algorithm2} outlines the scheduling and optimization process. After initialization, coalition scheduling is executed (lines 3-9), including dynamic estimation (line 4) and virtual queue computation (line 8). Based on these steps, the clients in $\pi(t)$ calculate $f_n^*$ to decide the computation resource (line 9). The primary complexity of Algorithm \ref{alg:algorithm2} stems from the sorting operations during coalition selection, with the time complexity of quicksort being $\mathit{O}(MlogM)$.

\subsection{Theoretical Analysis}
\paragraph{Boundary Analysis.} To satisfy participation balance, the scheduling rule must sometimes sacrifice the optimal choice of $g(t) := \left(1 - \frac{\hat{\mathcal{T}}_{m}(t)}{\mathcal{I}}\right)$ to maintain virtual queue stability. We use the Drift-plus-Penalty method to analyze this trade-off and derive bounds on the optimality gap.
\begin{theorem}
The time-average expected efficiency under the scheduling rule satisfies:
\begin{equation}
\frac{1}{\tau_g}\sum_{t=0}^{\tau_g-1}{\mathbb E}[g(t)] \ge g^*-{\cal O}(1/\beta) ,
\end{equation}
where $g^*$ is the optimal solution over all feasible options.
\end{theorem}
The result implies that a larger $\beta$ yields more near-optimal performance, but slower convergence and larger queue.
\paragraph{Convergence Analysis.} 
\begin{assumption}
The assumption parameters for convergence analysis are represented as a 4-tuple $(\sigma, \mu, \rho,[\backepsilon_1, \backepsilon_2, \backepsilon_3])$, where the function ${\cal F}_n(\omega)$ is $\sigma$-smooth, $\mu$-strong convexity, $\rho$-lipschitz \cite{assumption1} and gradient difference are bounded by $[\backepsilon_{1}, \backepsilon_{2}, \backepsilon_3]$ \cite{HFL_compute, HFL_drawback}.
\begin{itemize}
	\item $\sigma$-smooth, $\mu$-strong convexity, $\rho$-lipschitz: For $\forall \omega,\omega'$, $\left \| \nabla {\cal F}_n(\omega)- \nabla {\cal F}_n(\omega')\right \| \le \sigma\left \| \omega -\omega' \right \| $ ($\sigma >0$), $\left \langle \nabla {\cal F}_n(\omega'),\omega-\omega' \right \rangle +\frac{\mu}{2}\left \| \omega-\omega' \right \|^2 \le {\cal F}_n(\omega)-{\cal F}_n(\omega') $ ($\mu \ge 0$), $\left \| {\cal F}_n(\omega)-{\cal F}_n(\omega')  \right \|  \le \rho \left \| \omega-\omega' \right \|$.
	\item $\backepsilon_1$-bounded of gradient difference: The gradient difference of loss function between client $n$ and the associated ES $m$ is denoted as  $\left \| \nabla {\cal F}_n(\omega)-\nabla {\cal F}_m(\omega)  \right \|  \le \backepsilon_1$.

	\item $\backepsilon_2$-bounded of gradient difference: The gradient difference of loss function between ES $m$ and CS is denoted as  $\left \| \nabla {\cal F}_m(\omega)-\nabla {\cal F}(\omega)  \right \|^2  \le \backepsilon_2$, where $\backepsilon_2 = c_0+c_1\frac{M}{2}\overline{JS}$.
	
	\item $\backepsilon_3$-bounded of stochastic gradient: For $\forall n$, the expected squared norm of stochastic gradients is bounded by	${\mathbb E }\left \| \nabla F_n(\omega) \right \|^2 \le \backepsilon_3 $.
\end{itemize}
\label{convergence}
\end{assumption}
The convergence of SAFL is as follows:
\begin{theorem} 
	After $\tau_g$ global aggregations, SAFL convergences to a critical point, if the learning rate meets $\eta < \frac{1}{\sigma} $,
	\begin{equation}
		\begin{aligned}
			{\mathbb E}[{\cal F}(\omega(\tau_g)-{\cal F}(\omega^* ))]\le\eth[{\cal F}(\omega(0)-{\cal F}(\omega^* ))]\\
			+(1-\eth)\frac{O_1\backepsilon_3 +O_2\backepsilon   _1+O_3\backepsilon_2}{O_4},
		\end{aligned}
	\end{equation}
	where $\eth = (1-\xi_\varphi+\xi_\varphi(1-\eta\mu)^{\tau_c\tau_e})^{\tau_g}$, $O_1=\frac{1}{2\mu}$, $O_2=\rho\tau_e\frac{(\eta\sigma+1)^\tau_c-\sigma\eta\tau_c-1}{\sigma}$, $O_3=\frac{\tau_c\tau_e\eta}{2}$ ,and $O_4=1-(1-\eta\mu)^{\tau_c\tau_e}$.
	\label{convergence}
\end{theorem}
\paragraph{Remark:} As $\tau_g\to \infty $, $\eth \to 0$, for strongly convex functions, the convergence bound simplifies to $\frac{O_1\backepsilon_3 +O_2\backepsilon_1+O_3\backepsilon_2}{O_4}$, which is mainly influenced by $\backepsilon_1$, $\backepsilon_2$ and $\backepsilon_3$. The values of $\backepsilon_1$ and $\backepsilon_2$ depend on the edge association, highlighting the importance of adjusting the edge relationships to achieve a low $\overline{JS}$ in alignment with the intent of FedCure.

\section{Experiments}
\subsection{Experimental Configurations}
\paragraph{Datasets and Models.} We conduct experiments on four real-world datasets commonly used in FL: a CNN with 2 convolutional layers, 2 pooling layers and a fully connected layer on MNIST \cite{Mnist}; a CNN with 2 convolutional layers, one pooling layer and 3 fully connected layers on CIFAR-10 \cite{Cifar}, SVHN \cite{SVHN} and CINIC-10 \cite{Cinic}. 

\paragraph{Parameter Configurations.}The experiment involves 5 ESs and 50 clients. The initial clients’ data distributions are configured as edge non-IID following \cite{noniid_set}. The training process includes 5 rounds of local training, 12 rounds of edge iterations, and either 100 or 200 rounds of global iterations. The learning rate is set to 0.01 for MNIST, CIFAR-10 and CINIC-10, and 0.005 for SVHN. The other parameters are: $\ell=0.2$, $\Bbbk=[0.9, 0.99]$, and $\beta=0.5$.

\paragraph{Baselines.} 
Two clustering and two scheduling methods: {\bf K-Means} \cite{kmeans} is a clustering algorithm that segregates data into K clusters (K is pre-specified) and assigns data to the nearest cluster based on distance. {\bf Mean-Shift} \cite{meanshift} is a density-based non-parametric clustering algorithm that automatically determines the number of clusters. {\bf Greedy/FedGreedy} \cite{greedy} schedule coalitions by selecting options that maximize training efficiency. {\bf Fair/FedFair} \cite{SM3} leverages virtual queue to ensure balanced participation. 

\begin{figure}[t]
    \centering
    \begin{minipage}[b]{0.23\textwidth}
        \centering
        \includegraphics[scale=0.115]{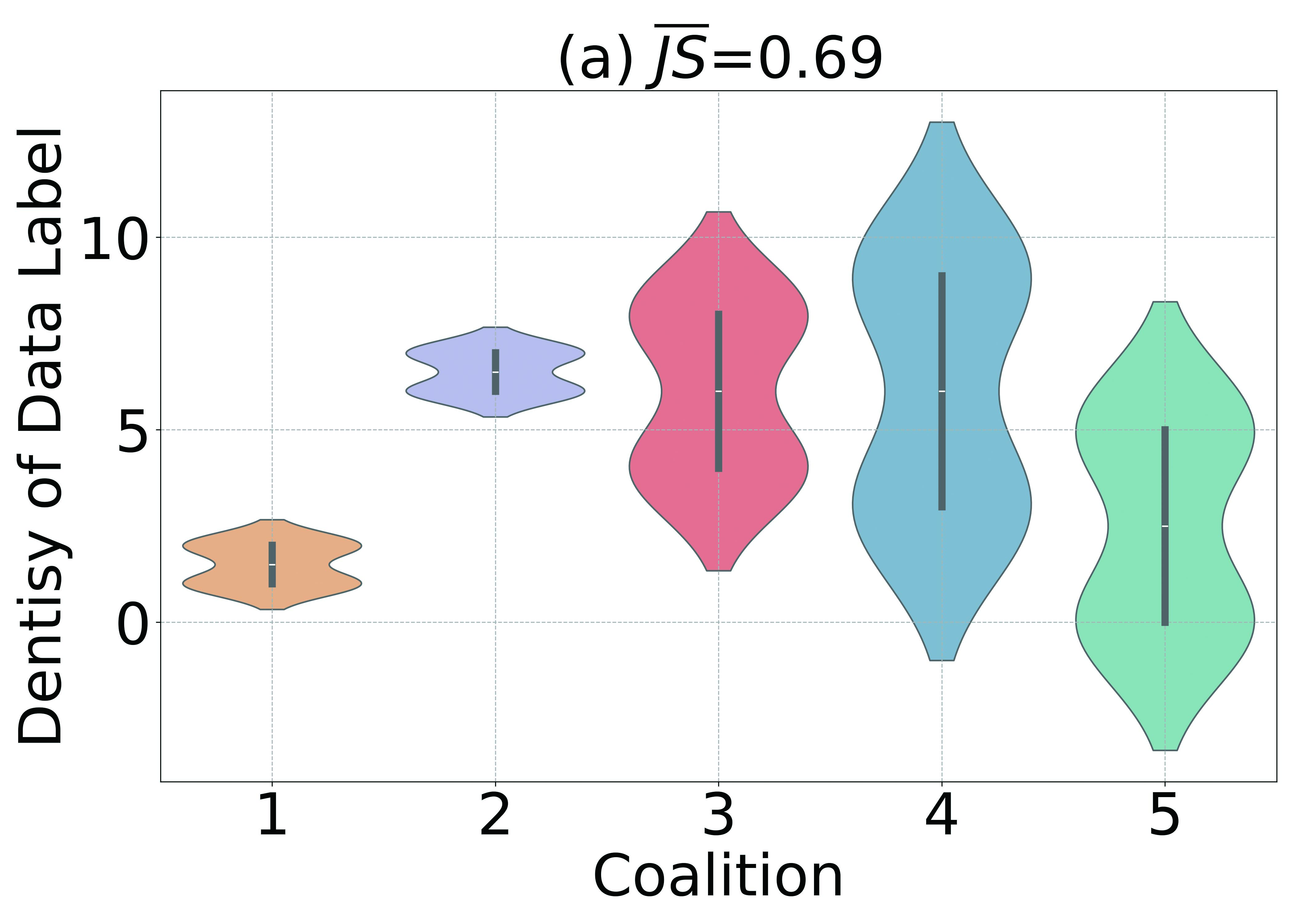}
    \end{minipage}
    \begin{minipage}[b]{0.23\textwidth}
        \centering
        \includegraphics[scale=0.115]{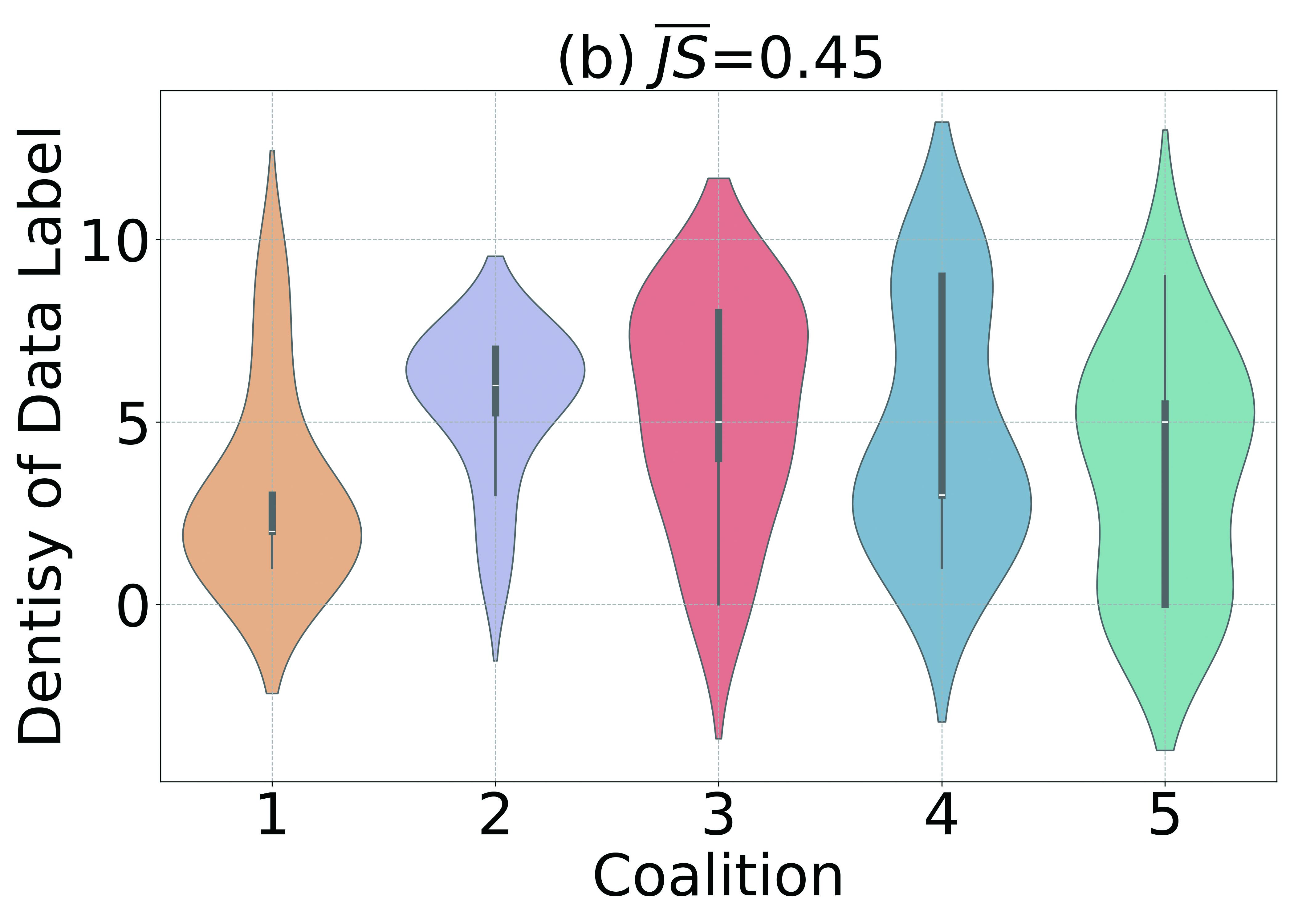}
    \end{minipage}
    \begin{minipage}[b]{0.23\textwidth}
        \centering
        \includegraphics[scale=0.115]{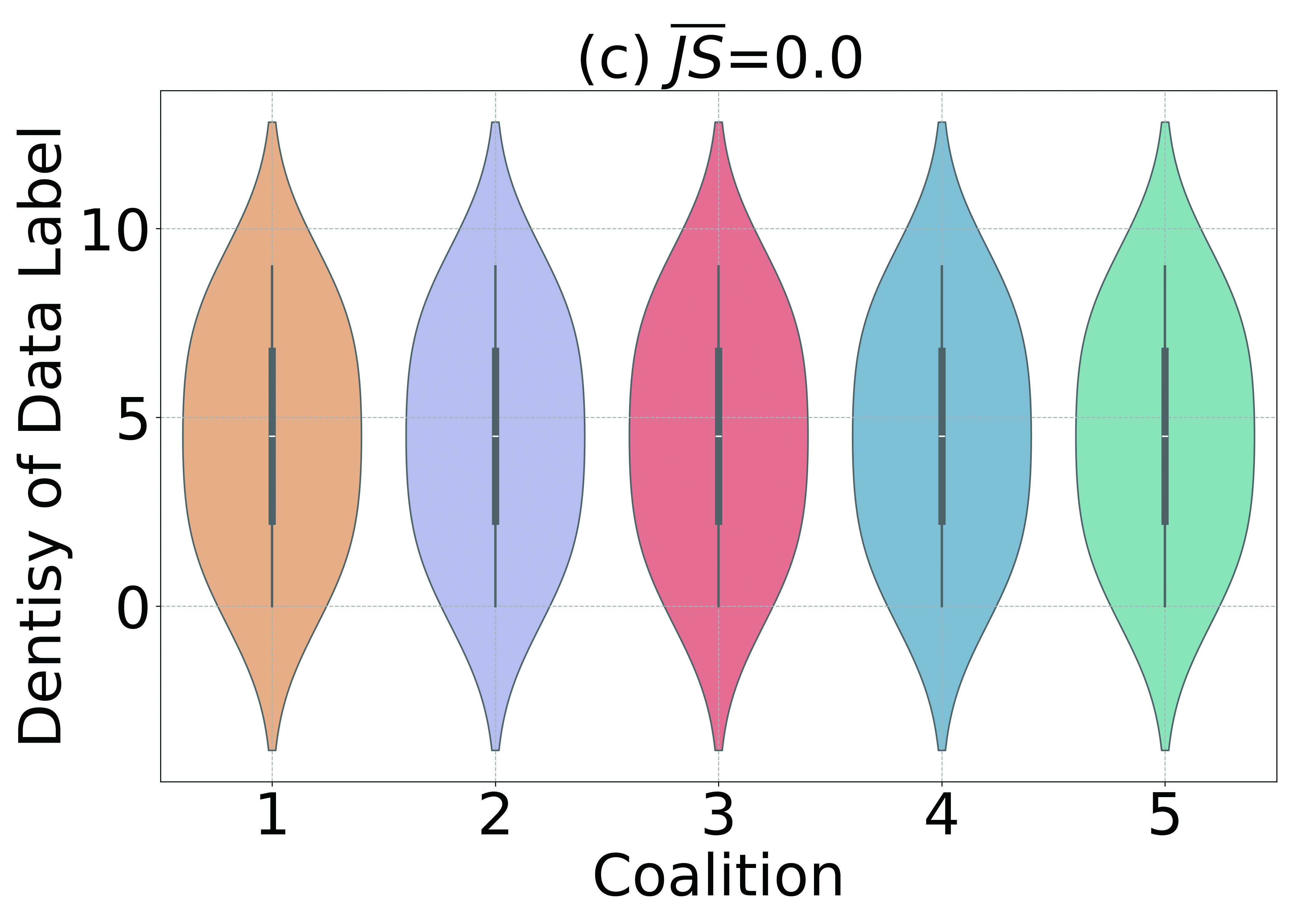}
    \end{minipage}
    \begin{minipage}[b]{0.23\textwidth}
        \centering
        \includegraphics[scale=0.115]{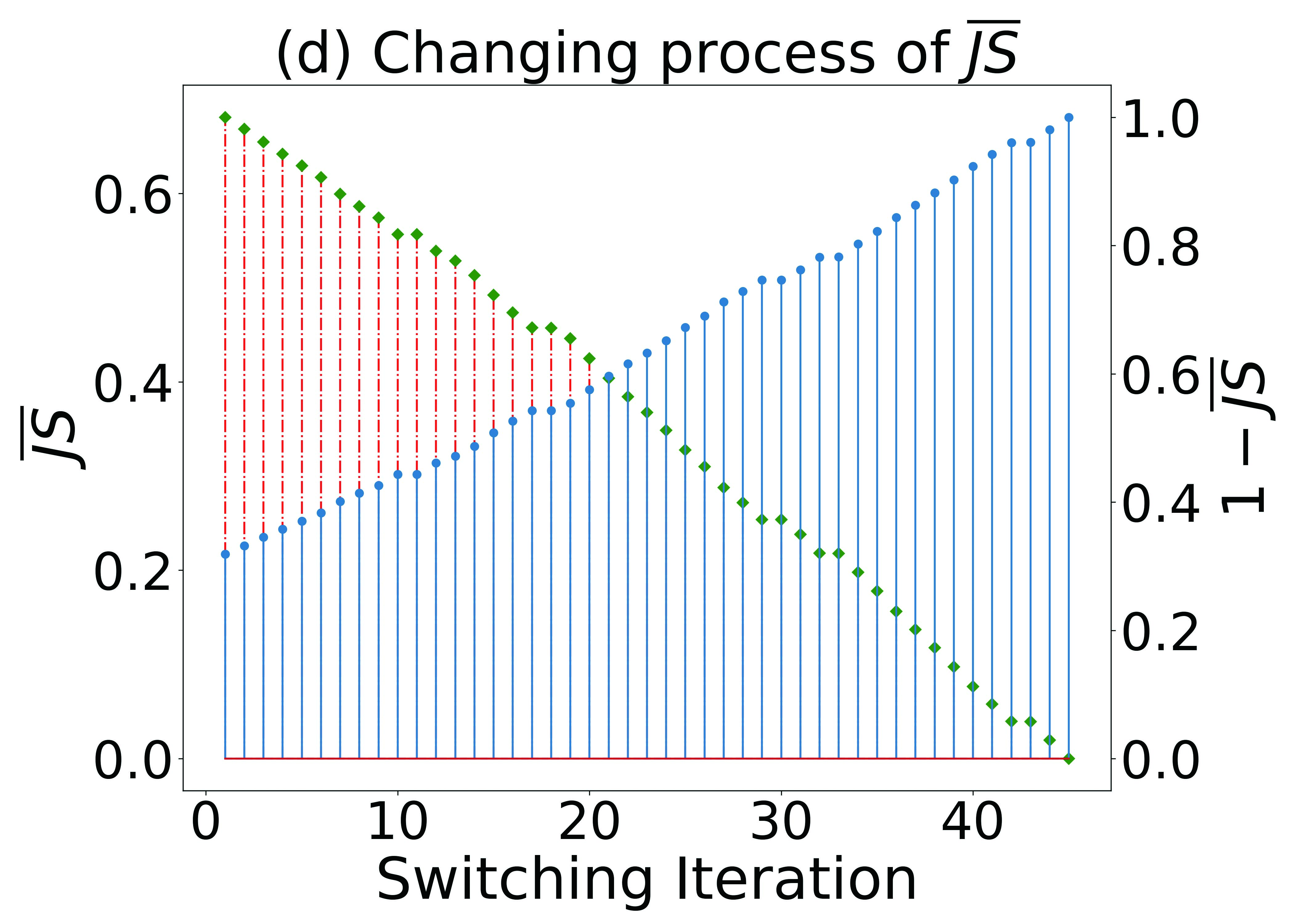}
    \end{minipage}
\caption{Data distribution variation and $\mathcal{\overline{JS}}$'s changing.}
\end{figure}

\subsection{Experimental Results}
\paragraph{Validating the effectiveness of FedCure on non-IID data.} Figures 2(a)-2(c) show the data distribution during the coalition formation process, with the width of the violin plots indicating the density of different data labels. The initial $\overline{JS}$ value is 0.69 and each coalition contains two label categories. Under the preference rule, the data distribution gradually becomes homogeneous, and the similarity of each coalition reaches its maximum in the final state where the $\overline{JS}$ value is 0. Figure 2(d) shows a consistent decline in $\overline{JS}$ value during client switching operations. Figure 3 shows the accuracy achieved using the two clustering algorithms (K-means and Mean-shift) and different data distributions. The maximum performance achieved based on the final data distributions is 1.40x, 3.09x, 2.95x, and 3.15x that of the initial state on four real-world datasets, respectively. The Mean-Shift algorithm includes five clusters, and we apply the same setting to K-means with $K = 5$. However, the repetition of labels in clusters limits the diversity of data distributions, and predefining $K$ will further hinder the applicability. Figures 2 and 3 validate that FedCure can achieve highly effective data distribution adjustments in terms of both accuracy and model convergence speed. This is evidenced by each shift towards lower $\overline{JS}$ values during coalition formation.

\setlength{\tabcolsep}{1mm} 
\begin{table}[t]
\centering
\begin{tabular}{>{\raggedright\arraybackslash}p{1.3cm} 
                >{\centering\arraybackslash}p{1.572cm} 
                >{\centering\arraybackslash}p{1.572cm} 
                >{\centering\arraybackslash}p{1.572cm} 
                >{\centering\arraybackslash}p{1.572cm}}  
\toprule  
\multirow{2}{*}{{\small Methods} } & \multicolumn{4}{c}{{\small Datasets}} \\   
\cmidrule{2-5}  
                          & {\small MNIST} & {\small CIFAR-10} & {\small SVHN} & {\small CINIC-10} \\   
\midrule  
 {\small Greedy}      &  {\small $37.97\pm0.03$} & {\small$18.25\pm0.06$} & {\small$9.36\pm1.08$} & {\small$14.56\pm1.84$} \\
 
{\small Fair}       & {\small $76.19\pm0.90$} & {\small$18.65\pm0.85$} & {\small$29.20\pm7.93$} & {\small$15.99\pm2.45$} \\

{\small FedGreedy}     & {\small $96.09\pm0.12$} & {\small$50.39\pm1.33$} & {\small$56.63\pm2.09$} & {\small$33.75\pm1.73$} \\

{\small FedFair}       & {\small $\underline{96.37}\pm0.05$} & {\small$\underline{53.15}\pm0.62$} & {\small$\underline{58.46}\pm1.13$} & {\small$34.83\pm1.72$} \\

{\small FedCure}     & {\small $\underline{96.37}\pm0.05$} & {\small$52.90\pm0.92$} & {\small$57.51\pm1.30$} & {\small$\underline{34.94}\pm1.66$} \\ 
\midrule
{\small $\text{Cohen's }d$} & {\small $0.00$} & {\small $0.32$} & {\small $0.78$} & {\small $0.07$} \\
\bottomrule 
\end{tabular} 
\caption{Accuracy comparison with baselines.}  
\label{table1}  
\end{table}

\begin{figure}[t]
    \centering
    \begin{minipage}[b]{0.23\textwidth}
        \centering
        \includegraphics[scale=0.116]{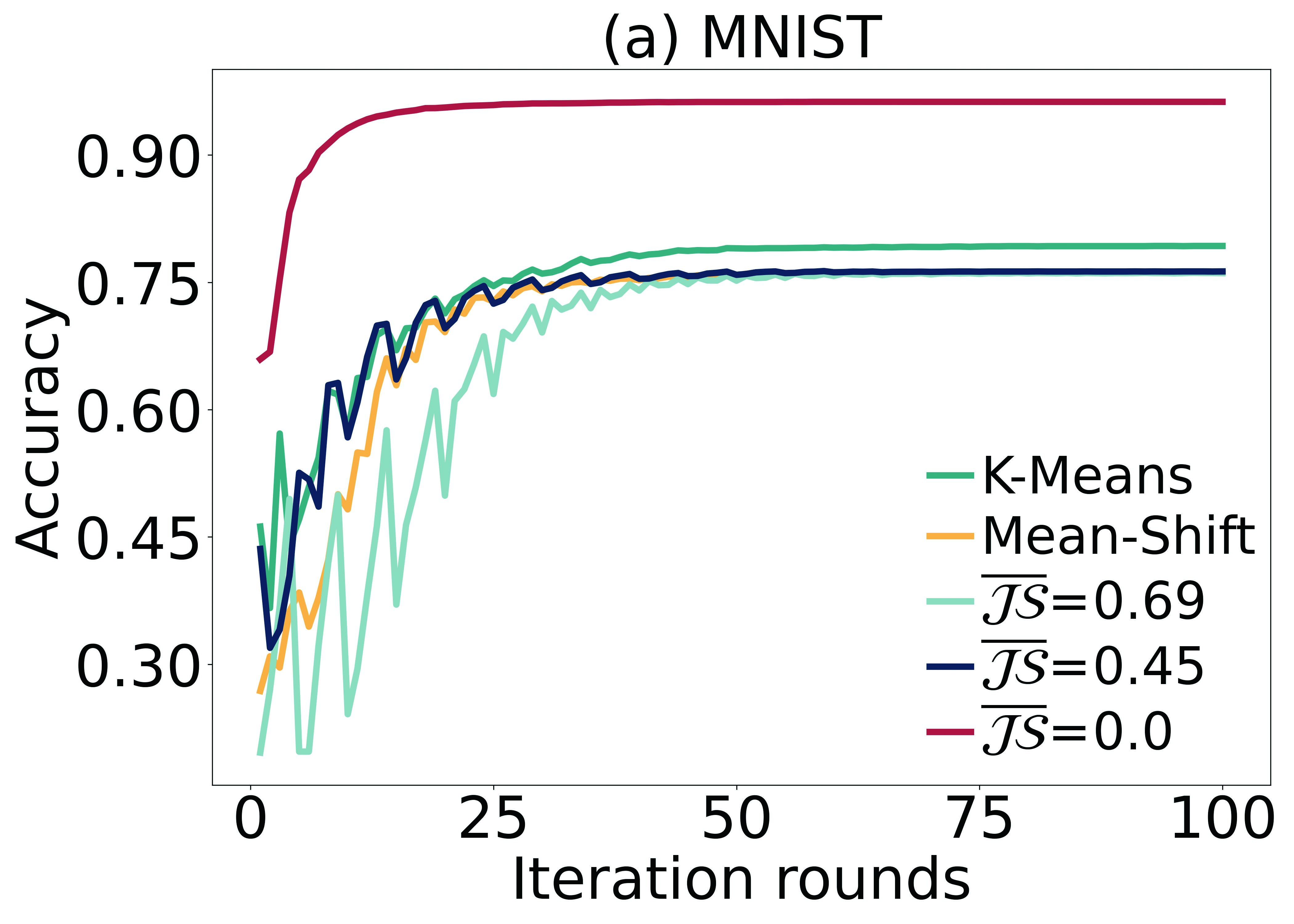}
    \end{minipage}
    \begin{minipage}[b]{0.23\textwidth}
        \centering
        \includegraphics[scale=0.116]{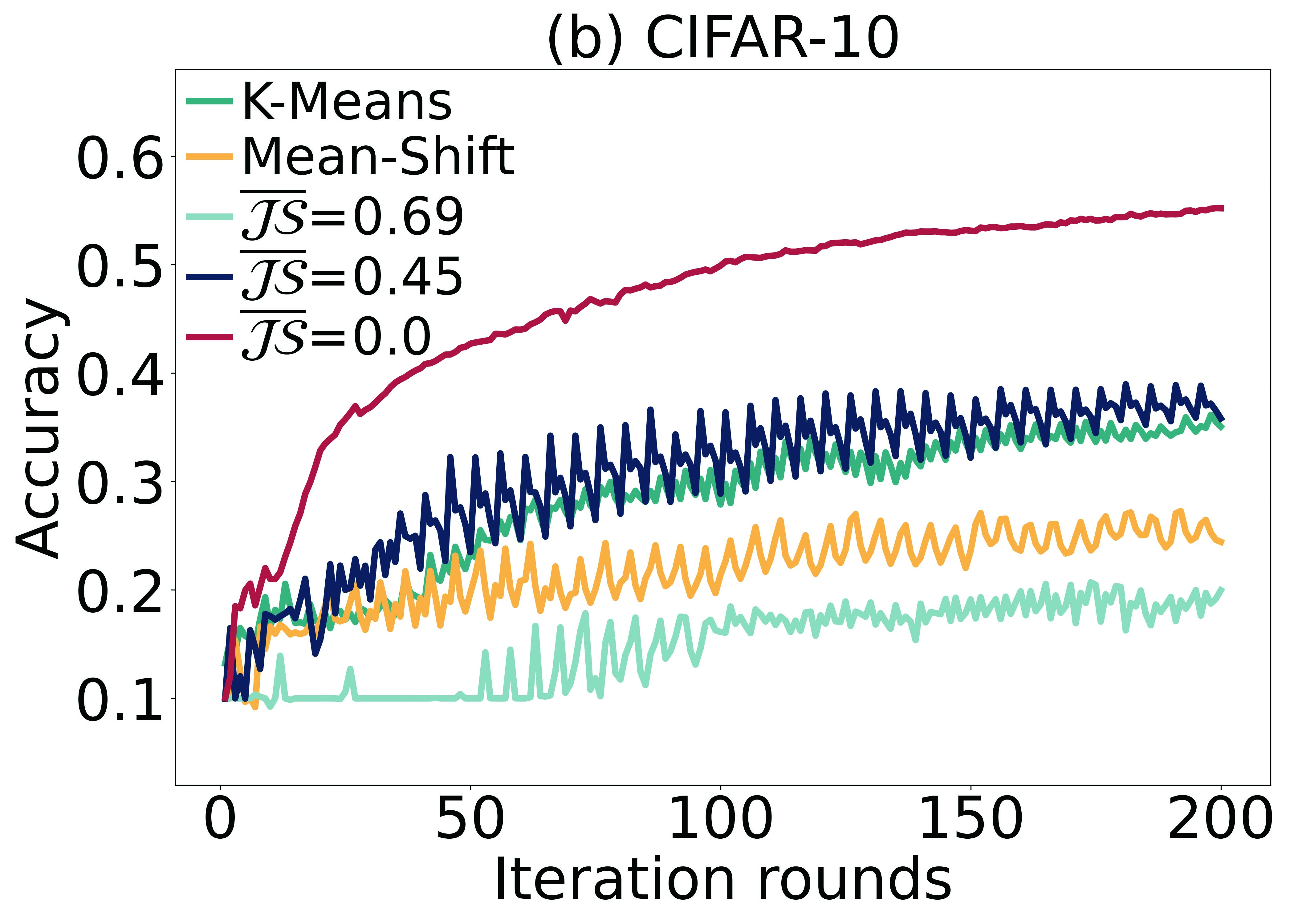}
    \end{minipage}
    \begin{minipage}[b]{0.23\textwidth}
        \centering
        \includegraphics[scale=0.116]{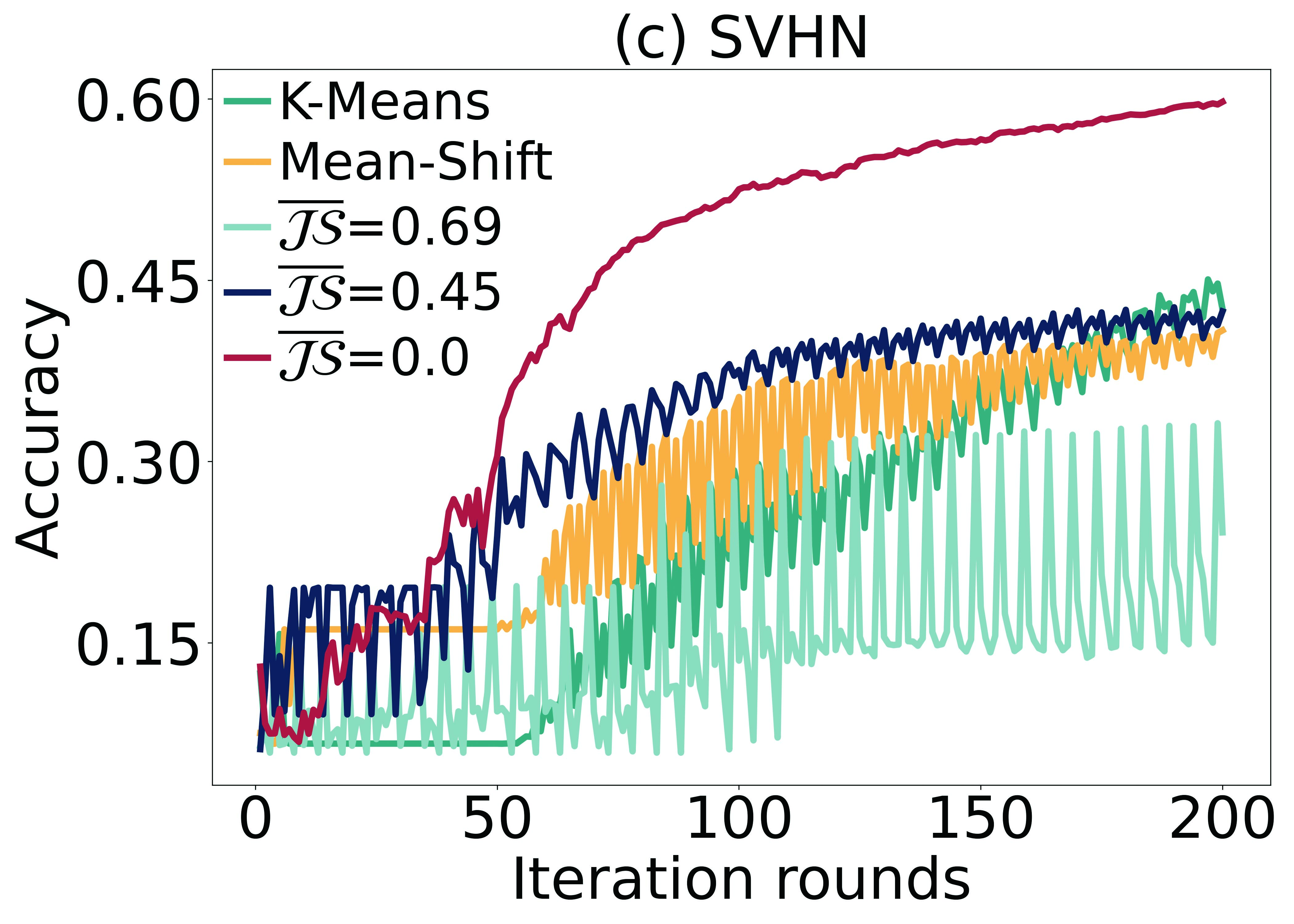}
    \end{minipage}
    \begin{minipage}[b]{0.23\textwidth}
        \centering
        \includegraphics[scale=0.116]{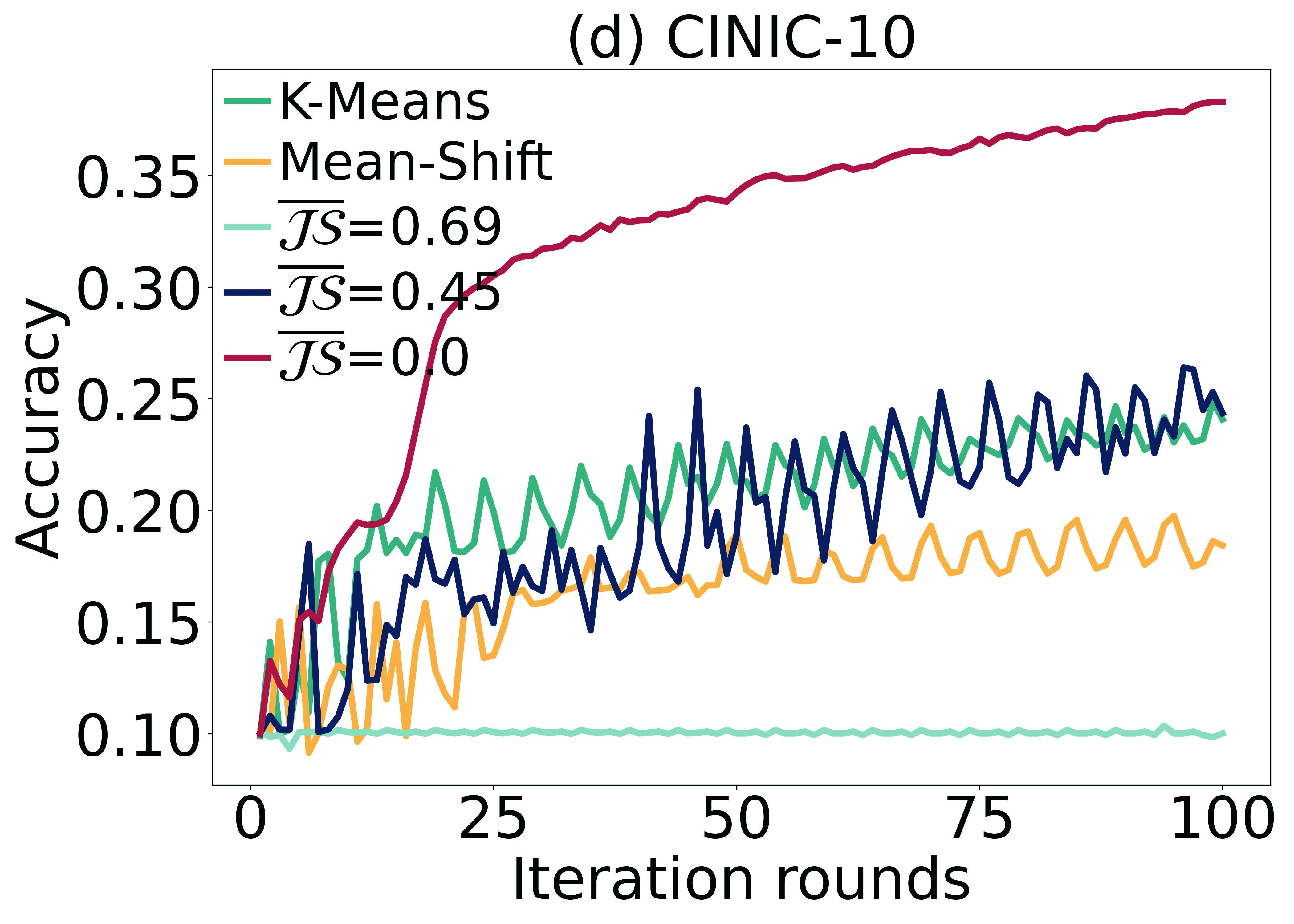}
    \end{minipage}
\caption{Comparison across clustering methods.}
\end{figure}

\paragraph{Validating the efficiency of scheduling and resource allocation.} 
Table~\ref{table1} summarizes the accuracy achieved by various scheduling methods. Greedy and Fair perform worse, mainly due to unadjusted data distribution. Greedy prioritizes coalitions with shorter training latency, exacerbating the issue of local optima. FedGreedy and FedFair, which rely on the stable coalitions provided by FedCure, achieve relatively higher accuracy. FedCure delivers accuracy comparable to FedFair, with the range of Cohen’s $d$ (0.00–0.78) indicating no substantial performance gaps, confirming that FedCure successfully sustains competitive accuracy through balance-based scheduling. Figure 4 presents the coefficient of variation (COV) of training latency per round, serving as a relative metric for comparing different scheduling methods. In Figure 4(a), FedCure and FedGreedy exhibit lower COVs, indicating comparable scheduling performance when prioritizing faster coalitions. Conversely, FedFair and Fair have higher COVs due to equal treatment of the fastest and slowest coalitions, resulting in greater relative variability. Combining Table \ref{table1} and Figure 4(a), FedCure effectively balances performance and efficiency. Figure 4(b) shows the impact of $\beta$ on virtual queue length. All curves stabilize after several rounds, verifying that FedCure maintains queue stability. Larger $\beta$ slows convergence and lengthens virtual queue, as they favor faster coalitions. While higher $\beta$ may compromise early balance, they still ensure long-term balance.
\begin{figure}[t]
    \centering
    \begin{minipage}[b]{0.5\textwidth}
        \centering
        \includegraphics[width=0.422\textwidth]{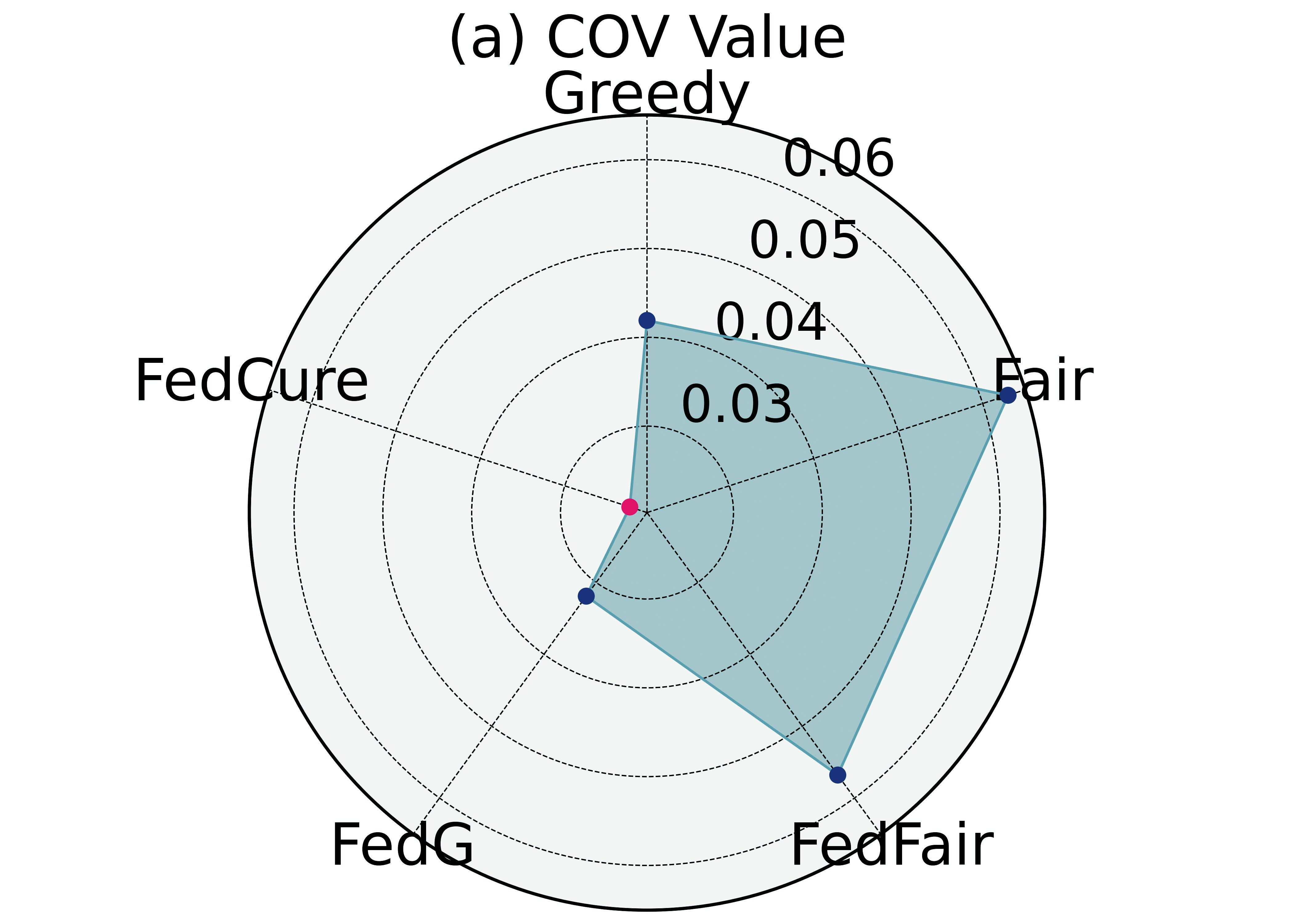}
        \includegraphics[width=0.425\textwidth]{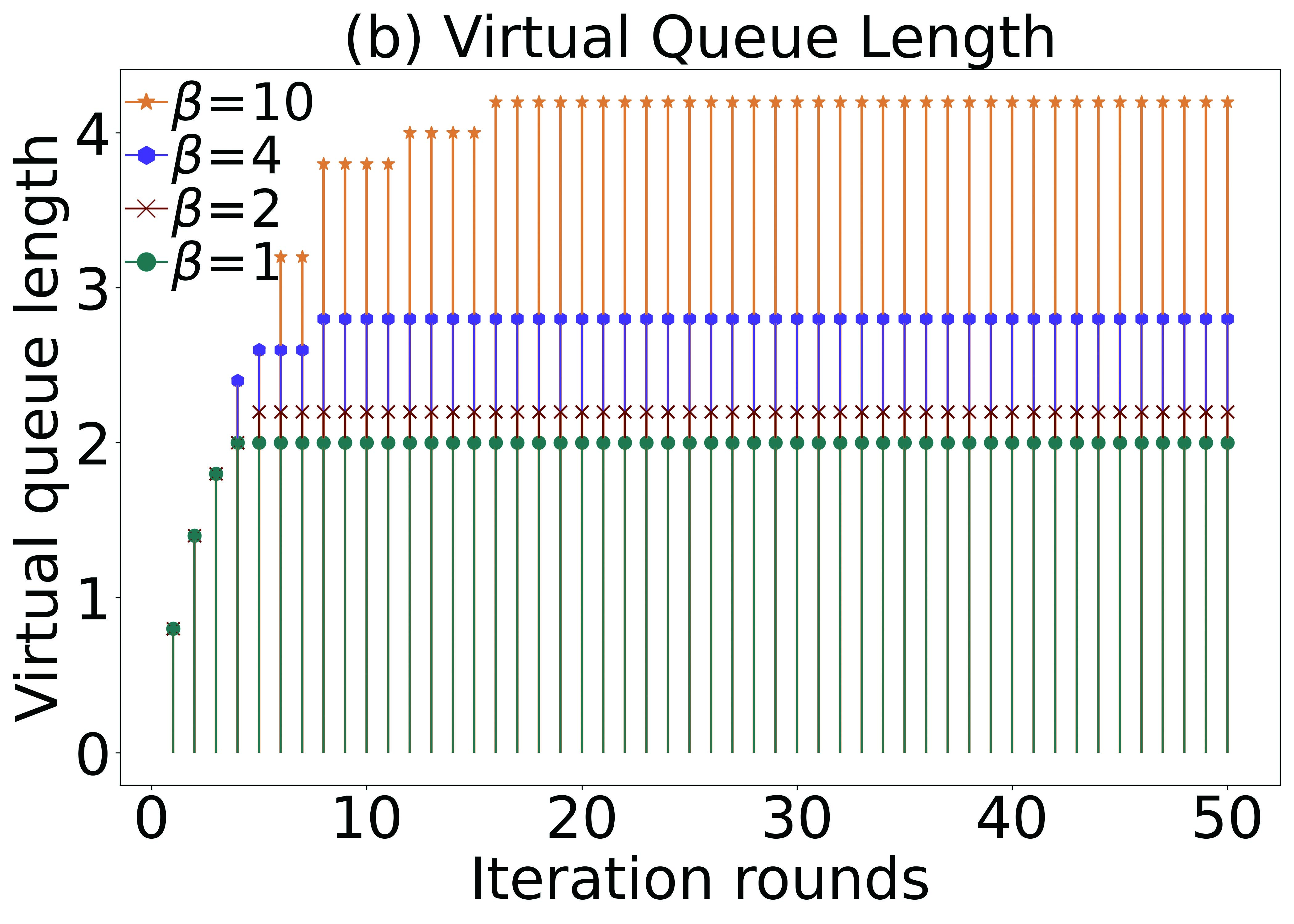}
        \caption{COV of training latency and queue length.}
    \end{minipage}
\end{figure}

\section{Conclusion}
In this paper, a novel framework, FedCure, was proposed to address the negative impacts of participation bias with non-IID data. Firstly, the data distribution in client-edge layer was transformed into a correlation problem, and a coalition formation game was established with a preference rule, reducing the non-IID degree among ESs. Secondly, the coalition scheduling in edge-cloud layer was guided by a scheduling rule that leverages Bayesian-estimated coalition dynamics and virtual queue, enhancing balance and efficiency. Based on this, a resource allocation rule was introduced to align with the estimated coalition dynamics by optimizing computation settings. Finally, extensive experiments on four real-world datasets validated FedCure's effectiveness. 


\section{Appendix}
\subsection*{Proof of Theorem 1}
The potential function $\phi$ of the coalition formation game is defined as the sum of $\overline{\mathcal{JS}}$ value among all coalitions. Mathematically, it is expressed as:
    \begin{equation}
        \phi \left ( {a_n} ,a_{-n} \right ) = \frac{1}{2}M(M-1) \overline{\mathcal{JS}}\left ( Q_{a_n},Q_{a_{-n}} \right ),
    \end{equation}
where $M$ represents the total number of coalitions, and $Q_{a_n}$ and $Q_{a_{-n}}$ denote the distributions associated with coalitions $a_n$ and $a_{-n}$, respectively.
When client $n$ switches its membership from coalition $a_n$ to $\widetilde{a_n}$, the corresponding change in the potential function is given by:
    \begin{equation}
    \begin{aligned}
         &  \phi \left ( \widetilde{a_n} ,a_{-n} \right ) -\phi \left ( a_n ,a_{-n} \right ) \\
        & =  \sum_{\underset{n \in \mathcal{G} _{\widetilde{a_n}}}{i=1,} }^{M-1} \sum_{j=i+1}^{M} \mathcal{JS}(Q_i,Q_j)  - \sum_{\underset{n \in \mathcal{G} _{a_n}}{i=1,} }^{M-1} \sum_{j=i+1}^{M} \mathcal{JS}(Q_i,Q_j)\\
        & = \sum_{j=\widetilde{a_n}+1 }^{M} \left [ \mathcal{JS}(Q_{\widetilde{a_n}\cup\left \{ n\right \}},Q_j)-\mathcal{JS}(Q_{\widetilde{a_n}\setminus \left \{ n\right \}},Q_j) \right]\\
        & \quad + \sum_{j=a_n+1 }^{M} \left [ \mathcal{JS}(Q_{a_n\setminus \left \{ n\right \}},Q_j)-\mathcal{JS}(Q_{a_n\cup  \left \{ n\right \}},Q_j) \right ]  \\
	   & \quad +\sum_{i=1}^{\widetilde{a_n}-1}\left[\mathcal{JS}(Q_i,Q_{\widetilde{a_n}\cup\left \{ n\right \}}) -\mathcal{JS}(Q_i,Q_{\widetilde{a_n}\setminus\left \{ n\right \}}) \right] \\
	   & \quad + \sum_{i=1}^{{a_n}-1}\left[ \mathcal{JS}(Q_i,Q_{{a_n}\setminus \left \{ n\right \}})- \mathcal{JS}(Q_i,Q_{{a_n}\cup \left \{ n\right \}}) \right]\\
        & \quad +\sum_{i\in \Omega}^{M-1} \sum_{\underset{j\in \Omega}{j=i+1,} }^{M} \left [ \mathcal{JS}(Q_i,Q_j)-\mathcal{JS}(Q_i,Q_j) \right ] ,
    \end{aligned}
    \end{equation}
    where $\Omega  $ denotes coalitions that are not relevant to this switching $\left \{ \mathcal{M}\setminus \left \{ a_n \right \} \setminus \left \{ \widetilde{a_n} \right \}   \right \} $. According to the preference rule $\Upsilon_p$, the JSD value for client $n$ joining a coalition is defined as:
    \begin{equation}
    \begin{aligned}
		& U_n(a_n ,a_{-n})\\
		&  =\sum_{j=a_n+1}^{M-1} \mathcal{JS}(Q_{a_n},Q_j) + \sum_{j=\widetilde{a_n} +1}^{M-1} JS(Q_{\widetilde{a_n} },Q_j) \\
		& \quad +\sum_{i=1}^{{a_n}-1}\mathcal{JS}(Q_i,Q_{{a_n}}) + \sum_{i=1}^{\widetilde{a_n}-1}\mathcal{JS}(Q_i,Q_{\widetilde{a_n}}) \\
		& \quad +\sum_{i\in \Omega  }^{M-1} \sum_{\underset{j\in \Omega}{j=i+1,}}^{M}\mathcal{JS}(Q_i,Q_j) .
    \end{aligned}
    \end{equation}
The difference in JSD values before and after the switch is derived as:
    \begin{equation}
    \begin{aligned}
         & U_n(\widetilde{a_n} ,a_{-n})-U_n(a_n ,a_{-n}) \\
         &=\sum_{j=\widetilde{a_n}  +1}^{M-1} \mathcal{JS}(Q_{\widetilde{a_n}\cup \left \{ n \right \}  },Q_j)+\sum_{j=a_n+1}^{M-1} \mathcal{JS}(Q_{a_n\setminus \left \{ n \right \} },Q_j)\\
         & \quad +\sum_{i=1}^{\widetilde{a_n}-1}\mathcal{JS}(Q_i,Q_{\widetilde{a_n}\cup\left \{ n\right \}}) + \sum_{i=1}^{{a_n}-1}\mathcal{JS}(Q_i,Q_{a_n\setminus \left \{ n \right \}})\\
         & \quad -\sum_{j=a_n+1}^{M-1} \mathcal{JS}(Q_{a_n\cup  \left \{ n \right \} },Q_j) - \sum_{j=\widetilde{a_n}+1}^{M-1} \mathcal{JS}(Q_{\widetilde{a_n}\setminus \left \{ n \right \} },Q_j)\\
	    & \quad -\sum_{i=1}^{{a_n}-1}\mathcal{JS}(Q_i,Q_{{a_n}\cup\left \{ n\right \}}) - \sum_{i=1}^{\widetilde{a_n}-1}\mathcal{JS}(Q_i,Q_{\widetilde{a_n}\setminus \left \{ n \right \}})\\
         & \quad + \sum_{i\in \Omega}^{M-1}\sum_{\underset{j\in \Omega}{j=i+1,} }^{M-1} \mathcal{JS}(Q_{i},Q_j)-\sum_{i\in \Omega  }^{M-1} \sum_{\underset{j\in \Omega}{j=i+1,} }^{M}\mathcal{JS}(Q_i,Q_j)\\
         & = \phi \left ( \widetilde{a_n} ,a_{-n} \right ) -\phi \left ( a_n ,a_{-n} \right ) .
    \end{aligned}
    \end{equation}
     Since the coalition formation game $\mathbb C$ is an exact potential game and satisfies the edge association constraint (EAC), it guarantees a stable partition. This completes the proof of Theorem 1.
     
\subsection{Proof of Theorem 2}
Given the estimated latency, stable coalition partition, and scheduled coalition ${\cal G}_{\pi(t)}$, each client within ${\cal G}_{\pi(t)}$ can optimize its computational resource allocation. We reformulate the original problem $\mathbb P$ as a conditional allocation problem ${\mathbb P}_1$:
\begin{subequations}
\begin{align}
{\mathbb{P}_1 }:& \quad \underset{f}{\max} \enspace \mathcal{Z}, \\
& \quad \text{s.t.}\quad f_n \in (0,f_n^{max}], \enspace  n\in {\cal G}_{\pi(t)}, \label{fmax}\\
& \quad \quad \quad \alpha>0,
\end{align}
\end{subequations}
where the objective function $\mathcal{Z}$ (as previously defined in ${\mathbb P}$) is represented as:
\begin{equation} 
	\mathcal{Z}=\mathbb{E} \left \{ \alpha (1-\frac{c_n}{f_n\hat{\mathcal{T}}_{\pi(t)}} ) -  \gamma (f_n )^ \varsigma \right \}, m\in{\pi(t)}.
	\label{Z}
\end{equation}
To determine the optimal resource allocation $f_{n}^*$, we analyze the properties of $\mathcal{Z}$:
\begin{numcases}{}
	\frac{\partial \mathcal{Z} }{\partial f_n} = \frac{\alpha c_n}{f_n^2 \hat{\mathcal{T}}_{\pi(t)} } -\varsigma \gamma (f_n)^{\varsigma -1},
	\label{first_derivative}\\
	\frac{\partial^2 \mathcal{Z} }{\partial f_n^2} = -\frac{2\alpha c_n}{f_n^3 \hat{\mathcal{T}}_{\pi(t)} } -\varsigma(\varsigma-1) \gamma (f_n)^{\varsigma -2} < 0.
\end{numcases}
The negativity of the second derivative confirms that $\cal Z$ is a strictly concave function with respect to $f_n$. By setting the first derivative to zero, we obtain the candidate solution:
\begin{equation}
	c = \sqrt[\varsigma +1]{\frac{\alpha c_n}{\varsigma \gamma \hat{\mathcal{T}}_{\pi(t)}} }.
\end{equation} 
Incorporating the restriction of $f_{n}$ in Eq. (\ref{fmax}), the optimal resource allocation $f_{n}^*$ is
\begin{equation}
    f_{n}^*=\min\left \{{f_n^{max}} ,\sqrt[\varsigma +1]{\frac{\alpha c_n}{\varsigma \gamma \hat{\mathcal{T}}_{\pi(t)}} }\right \}.
\end{equation}
This completes the proof of Theorem 2.

\subsection*{Proof of Theorem 3}
To establish Theorem 3, we first present a key lemma inspired by Theorem 4.5 in \cite{queue}.
\begin{lemma}
There exists a $\Theta$-only policy $\vartheta$ that makes independent and stationary decisions as a pure (possibly randomized) function of the observed $\Theta(t)$ for each round and for any $\varepsilon$,
	\begin{numcases}{}
		g^{s}-{\mathbb E}[ {\textstyle \sum_{m=1}^{M}} \chi_m^\vartheta(t)(1 - \frac{\hat{\mathcal{T}}_m(t)}{\cal I})]\le \varepsilon,\\
		\delta _m\le {\mathbb E}[\chi_m^\vartheta(t)]+\varepsilon ,\forall m\in \cal M,
	\end{numcases}
where the expectation is taken with respect to the random decisions made under the policy $\vartheta$ and the stochastic event $\Theta(t)$, and $g^{s}$ represents the supremum value of ${\mathbb E}[{\textstyle \sum_{m=1}^{M}} \chi_m^\vartheta(t)(1 - \frac{\hat{\mathcal{T}}_m(t)}{\cal I})]$ across all potential policies.
\end{lemma}
While $\Theta(t)$ in FedCure is non-stationary due to asynchronous coalition selection, its values are drawn from a finite set. This property ensures the continued applicability of Lemma 1. We now establish the stability proof as follows:

First, we define the Lyapunov function:
\begin{equation}
	\mathcal{L}(\mathbf{\Lambda}(t))  =\frac{1}{2}  {\textstyle \sum_{m=1}^{M}} \Lambda_m^2(t),
\end{equation}
where $\mathbf{\Lambda}(t)=[\Lambda_1 (t),\cdots ,\Lambda_M (t)]$ denotes the queue length vector at round $t$. Then, we have the drift of the Lyapunov function:
\begin{equation}
\begin{aligned}
& \mathcal{L}(\mathbf{\Lambda}(t+1))-\mathcal{L}(\mathbf{\Lambda}(t))\\
& \le \frac{1}{2} \sum_{m=1}^{M} [\Lambda _m(t)+\delta _m-\chi_m(t)]^2-\frac{1}{2} \sum_{m=1}^{M} [\Lambda _m(t)]^2 \\
& \le \frac{K}{2} + \sum_{m=1}^{M}\delta _m\Lambda _m(t)-\sum_{m=1}^{M}\chi _m(t)\Lambda _m(t),
\end{aligned}
\end{equation}
where $K=\sum_{m=1}^{M}\left [ \delta _m- \chi_m(t)\right ] ^2 \le 1$ represents the bounded variance term. Next, by applying conditional expectation, we obtain
\begin{equation}
\begin{aligned}
& \mathbb{E}[ \mathcal{L}(\mathbf{\Lambda}(t+1))-\mathcal{L}(\mathbf{\Lambda}(t))|\mathbf{\Lambda}(t)]\\
& \le \frac{K}{2} + \sum_{m=1}^{M}\delta _m\Lambda _m(t)- \mathbb{E}[\sum_{m=1}^{M}\chi _m(t)\Lambda _m(t)|\mathbf{\Lambda}(t)]\\
& \le R +   \sum_{m=1}^{M}\delta _m\Lambda _m(t) \\
& \quad - \mathbb{E}\left \{\sum_{m=1}^{M}\chi _m(t)[\Lambda_m (t) + \beta ( 1 - \frac{\hat{\mathcal{T}}_m(t)}{\cal I}) ]|\mathbf{\Lambda}(t)\right\},
\end{aligned}
\label{con_exp}
\end{equation}
where $R=\frac{K}{2}+\beta$. The lower bound for the last term in the RHS of Eq. (\ref{con_exp}) can be derived as:
\begin{equation}
\begin{aligned}
& \mathbb{E}\left \{\sum_{m=1}^{M}\chi _m(t)[\Lambda_m (t) + \beta ( 1 - \frac{\hat{\mathcal{T}}_m(t)}{\cal I}) ]|\mathbf{\Lambda}(t)\right\}\\
& \ge \mathbb{E}\left \{\sum_{m=1}^{M}\chi_m^{\vartheta}(t)[\Lambda_m (t) + \beta ( 1 - \frac{\hat{\mathcal{T}}_m(t)}{\cal I}) ]|\mathbf{\Lambda}(t)\right\}\\
& \ge \mathbb{E}\left \{\sum_{m=1}^{M}\chi_m^{\vartheta}(t)\Lambda_m (t)|\mathbf{\Lambda}(t)\right\}\\
& \ge \sum_{m=1}^{M}(\delta _m-\varepsilon )\Lambda _m(t),
\end{aligned}
\label{con_exp_lowerbound}
\end{equation}
where the inequality holds for no other $\Theta$-only policies could match up with the scheduling rule of FedCure in finding the biggest element in $\left \{ \Lambda_m (t) + \beta ( 1 - \frac{\hat{\mathcal{T}}_m(t)}{\cal I} )\right \}$, $m\in \cal M $. Let $\varepsilon \to 0$, we have the following inequality:
\begin{equation}
\begin{aligned}
& \mathbb{E}\left \{\sum_{m=1}^{M}\chi _m(t)[\Lambda_m (t) + \beta ( 1 - \frac{\hat{\mathcal{T}}_m(t)}{\cal I}) ]|\mathbf{\Lambda}(t)\right\} \\
& \ge \sum_{m=1}^{M}\delta _m\Lambda _m(t).
\end{aligned}
\label{to_0}
\end{equation}
By substituting Eq. (\ref{to_0}) into Eq. (\ref{con_exp}) and summing over all rounds $t \in \left\{1,\cdots, \tau_g \right\}$ as $\tau_g \to \infty $ simultaneously, we obtain the key bound:
\begin{equation}
	\mathbb{E}[ \mathcal{L}(\mathbf{\Lambda}(\tau_g))-\mathcal{L}(\mathbf{\Lambda}(0))] \le \tau_g R.
\end{equation}
Applying the following properties that ${\mathbb E}[\Lambda_m(t)]^2 \le{\mathbb E}[\Lambda_m^2(t)] $ and ${\mathbb L}(\Lambda(0))=0$, we derive the quadratic bound:
\begin{equation}
	\sum_{m=1}^{M} {\mathbb E}[\Lambda_m(t)]^2 \le  \sum_{m=1}^{M} {\mathbb E}[\Lambda^2_m(t)] \le 2\tau_g R.
\end{equation}
Through application of  Jensen's inequality, this yields the asymptotic result:
\begin{equation}
	\underset{\tau_g \to \infty}{\lim} \sum_{m=1}^{M} \frac{{\mathbb E}[\Lambda_m(\tau_g)]}{\tau_g} \le \underset{\tau_g \to \infty}{\lim} \sqrt{\frac{2MR}{\tau_g}}=0.
\end{equation}
Finally, we conclude that $\underset{\tau_g \to \infty}{\lim} \frac{{\mathbb E}[\Lambda_m(\tau_g)]}{\tau_g}=0, \forall m \in \cal M$. This completes the proof of Theorem 3.

\subsection*{Proof of Theorem 4}
Building upon Theorem 3's proof, we first establish the conditional Lyapunov drift: 
\begin{equation}
\Delta(t) =\mathbb{E}[ \mathcal{L}(\mathbf{\Lambda}(t+1))-\mathcal{L}(\mathbf{\Lambda}(t))|\mathbf{\Lambda}(t)].
\end{equation}
This satisfies the fundamental inequality:
\begin{equation}
 \Delta(t) \le \frac{K}{2} + \sum_{m=1}^{M}\delta _m\Lambda _m(t)- \mathbb{E}[\sum_{m=1}^{M}\chi _m(t)\Lambda _m(t)|\mathbf{\Lambda}(t)].
\end{equation}
Let $g(t) := \left(1 - \frac{\hat{\mathcal{T}}_{m}(t)}{\mathcal{I}}\right)$ be the penalty function, we then construct the drift-plus-penalty expression as follows:
\begin{equation}
\begin{aligned}
&\Delta(t) - \beta \cdot \mathbb{E}[g(t)] \\
& \le \frac{K}{2} + \sum_{m=1}^{M}\Lambda _m(t)(\delta _m- \mathbb{E}[\chi _m(t)|\mathbf{\Lambda}(t)])- \beta \cdot \mathbb{E}[g(t)].
\end{aligned}
\end{equation}
Let $g^*$ denote the optimal long-term average utility subject to all system constraints, from which it follows that 
\begin{equation}
\sum_{m=1}^{M}\Lambda _m(t)(\delta _m- \mathbb{E}[\chi _m(t)|\mathbf{\Lambda}(t)]) \ge 0.
\end{equation}
It simplifies the drift-plus-penalty expression to:
\begin{equation}
\Delta(t) - \beta \cdot \mathbb{E}[g(t)] \le \frac{K}{2} - \beta g^*.
\label{delta}
\end{equation}
By summing both sides of this inequality over $t = 0$ to $\tau_g - 1$, we obtain
\begin{equation}
\begin{aligned}
&\sum_{t=0}^{\tau_g-1} \mathbb{E}[ \mathcal{L}(\mathbf{\Lambda}(t+1))-\mathcal{L}(\mathbf{\Lambda}(t))]-  \beta \sum_{t=0}^{\tau_g-1}\mathbb{E}[g(t)]\\
& = \mathcal{L}(\mathbf{\Lambda}(\tau_g)) -\mathcal{L}(\mathbf{\Lambda}(0)) - \beta \sum_{t=0}^{\tau_g-1}\mathbb{E}[g(t)]\\
& \le \frac{\tau_g K}{2} -\tau_g \beta g^*.
\end{aligned}
\label{L0}
\end{equation}
Since the Lyapunov function $\mathcal{L}(\mathbf{\Lambda}(\cdot))\ge 0$ and $\mathcal{L}(\mathbf{\Lambda}(0))=0$, we have
\begin{equation}
- \beta \sum_{t=0}^{\tau_g-1}\mathbb{E}[g(t)] \le \frac{\tau_g K}{2} -\tau_g \beta g^*.
\end{equation}
Dividing both sides by $\tau_g\beta$ yields
\begin{equation}
\frac{1}{\tau_g}{\sum_{t=0}^{\tau_g-1}\mathbb{E}[g(t)]} \ge g^* - \frac{K}{2\beta}.
\end{equation}
This can be equivalently expressed as
\begin{equation}
\frac{1}{\tau_g}\sum_{t=0}^{\tau_g-1}{\mathbb E}[g(t)] \ge g^*-{\cal O}(1/\beta).
\end{equation}
Hence, Theorem 4 is proved.

\subsection*{Proof of Theorem 5}
Building upon prior work \cite{conv_1}, we focus on the distinctive aspects of our convergence proof. The CS satisfies the following key inequality:
\begin{equation}
	\begin{aligned}
& \mathbb{E}[{\cal F}(\omega^t)-{\cal F}(\omega^*)]\\
& \le ( 1-\xi_{\varphi}) [{\cal F}(\omega^{t-1})-{\cal F}(\omega^*)]+\xi_{\varphi}{\mathbb E}[{\cal F}(\omega_m^{t_\tau})-{\cal F}(\omega^*)]\\
& \overset{(a)}{\le}   ( 1-\xi_{\varphi}) [{\cal F}(\omega^{t-1})-{\cal F}(\omega^*)] + \xi_{\varphi}[(1-\eta\mu )^{\tau_c\tau_e}\\
& \quad \quad [\mathcal{F}(\omega^{t_\tau})-\mathcal{F}(\omega^*)]+\rho  \tau_e\mathcal{H} (\tau_c)+\frac{\tau_c\tau_e\eta\backepsilon_2}{2}]\\
& =  ( 1-\xi_{\varphi}+ \xi_{\varphi}(1-\eta\mu )^{\tau_c\tau_e} )[{\cal F}(\omega^{t-1})-{\cal F}(\omega^*)]+\\
& \quad \quad \xi_{\varphi}(1-\eta\mu )^{\tau_c\tau_e}[\mathcal{F}(\omega^{t_\tau})-\mathcal{F}(\omega^{t-1})]+\\
& \quad \quad \xi_{\varphi}(\rho  \tau_e\mathcal{H} (\tau_c)+\frac{\tau_c\tau_e\eta\backepsilon_2}{2}),
\end{aligned}
\label{conv_1}
\end{equation}
where $(a)$ holds because the upper bound between the edge model and the optimal global model can be derived as \cite{conv_3}:
\begin{equation}
	\begin{aligned}
	& \mathcal{F}(\omega_m^{t_\tau})-\mathcal{F}(\omega^*) \\
	& \le (1-\eta\mu )^{\tau_c\tau_e}[\mathcal{F}(\omega^{t_\tau})-\mathcal{F}(\omega^*)]+\rho  \tau_e\mathcal{H} (\tau_c)+\\
	& \qquad \frac{\tau_c\tau_e\eta\backepsilon_2}{2},
	\end{aligned}\\
\end{equation}
where $\mathcal{H}(x) \overset{\triangle }{=}  \frac{\backepsilon_1}{\sigma}((\eta\sigma+1)^x-1)-\eta\sigma x$.
For analytical tractability, we define the RHS of Eq. (\ref{conv_1}) as:
\begin{numcases}{}
   U=1-\xi_{\varphi}+ \xi_{\varphi}(1-\eta\mu )^{\tau_c\tau_e},\label{conv_U}\\
   U_1=U\cdot[{\cal F}(\omega^{t-1})-{\cal F}(\omega^*)], \label{conv_U1}\\
   U_2=\xi_{\varphi}(\rho  \tau_e\mathcal{H} (\tau_c)+\frac{\tau_c\tau_e\eta\backepsilon_2}{2}).\label{conv_K}
\end{numcases}
Substituting Eqs. (\ref{conv_U})-(\ref{conv_K}) into the RHS of Eq. (\ref{conv_1}) yields:
\begin{equation}
	\begin{aligned}
	& \mathbb{E}[{\cal F}(\omega^t)-{\cal F}(\omega^*)]\\
& \le \xi_{\varphi}(1-\eta\mu )^{\tau_c\tau_e}[\mathcal{F}(\omega^{t_\tau})-\mathcal{F}(\omega^{t-1})]+U_1+U_2\\
& \le \xi_{\varphi}(1-\eta\mu )^{\tau_c\tau_e}[\mathcal{F}(\omega^{t_\tau})-\mathcal{F}(\omega^*)]+U_1+U_2\\
& \le \xi_{\varphi}(1-\eta\mu )^{\tau_c\tau_e}\frac{1}{2\mu}\left \| \nabla{\cal F}(\omega^{t_\tau})\right \|^2+U_1+U_2\\
& \le \xi_{\varphi}(1-\eta\mu )^{\tau_c\tau_e}\frac{\backepsilon_3}{2\mu}+U_1+U_2\\
& \le \xi_{\varphi}(\frac{\backepsilon_3}{2\mu}+\rho  \tau_e\mathcal{H} (\tau_c)+\frac{\tau_c\tau_e\eta\backepsilon_2}{2})+U_1.
\end{aligned}
\label{conv_2}
\end{equation}
By telescoping and taking total expectation, after $\tau_g$ global updates on the CS, we have
\begin{equation}
\begin{aligned}
& \mathbb{E}[{\cal F}(\omega^{\tau_g})-{\cal F}(\omega^*)]\\
& \le U^{\tau_g}[{\cal F}(\omega^0)-{\cal F}(\omega^*)]+\\
& \quad \quad \xi_{\varphi}(\frac{\backepsilon_3}{2\mu}+\rho  \tau_e\mathcal{H} (\tau_c)+\frac{\tau_c\tau_e\eta\backepsilon_2}{2})\sum_{i=1}^{\tau_g} U^{i-1}.
\end{aligned}
\label{conv_3}
\end{equation}
By setting $U_3=\frac{\backepsilon_3}{2\mu}+\rho  \tau_e\mathcal{H} (\tau_c)+\frac{\tau_c\tau_e\eta\backepsilon_2}{2}$, so that Eq. (\ref{conv_3}) can be simplified as
\begin{equation}
\begin{aligned}
&\mathbb{E}[{\cal F}(\omega^{\tau_g})-{\cal F}(\omega^*)]\\
& \le U^{\tau_g}[{\cal F}(\omega^0)-{\cal F}(\omega^*)]+\xi_{\varphi}U_3\sum_{i = 1}^{\tau_g} U^{i-1}\\
& = U^{\tau_g}[{\cal F}(\omega^0)-{\cal F}(\omega^*)]+\xi_{\varphi}U_3\frac{1-U^{\tau_g}}{\xi_{\varphi}(1-U)}.
\end{aligned}
\label{conv_4}
\end{equation}
Next, we give the full results
\begin{equation}
\begin{aligned}
&\mathbb{E}[{\cal F}(\omega^{\tau_g})-{\cal F}(\omega^*)]\\
& \le [1-\xi_{\varphi}+ \xi_{\varphi}(1-\eta\mu )^{\tau_c\tau_e}]^{\tau_g}[{\cal F}(\omega^0)-{\cal F}(\omega^*)]+\\
& \quad \quad [\frac{\backepsilon_3}{2\mu}+\rho  \tau_e\mathcal{H} (\tau_c)+\frac{\tau_c\tau_e\eta\backepsilon_2}{2}]\cdot\\
& \quad \quad \frac{1-[1-\xi_{\varphi}+ (1-\eta\mu )^{\tau_c\tau_e}]^{\tau_g}}{1-(1-\eta\mu )^{\tau_c\tau_e}}.
\end{aligned}
\label{conv_5}
\end{equation}
Rearranging Eq. (\ref{conv_5}), we get
\begin{equation}
\begin{aligned}
&\mathbb{E}[{\cal F}(\omega^{\tau_g})-{\cal F}(\omega^*)]  \le \eth[{\cal F}(\omega(0)-{\cal F}(\omega^* ))]\\
& \qquad +(1-\eth)\frac{O_1\backepsilon_3 +O_2\backepsilon_1+O_3\backepsilon_2}{O_4},
\end{aligned}
\label{conv_6}
\end{equation}
where
\begin{numcases}{}
   \eth = (1-\xi_\varphi+\xi_\varphi(1-\eta\mu)^{\tau_c\tau_e})^{\tau_g},\\
   O_1=\frac{1}{2\mu},\\
   O_2=\rho\tau_e\frac{(\eta\sigma+1)^\tau_c-\sigma\eta\tau_c-1}{\sigma},\\
   O_3=\frac{\tau_c\tau_e\eta}{2},\\
   O_4=1-(1-\eta\mu)^{\tau_c\tau_e}.
\end{numcases}
Consequently, Theorem 5 is proved.

\subsection*{Supplement of Experiments}
To further substantiate FedCure's effectiveness beyond the main text's scope, we present additional experiments comparing FedCure's performance with RH \cite{RH}, a reputation-aware hedonic coalition formation algorithm employing selfish preference rules. In RH, clients form stable coalitions based on cluster head reputation and individual utility maximization, exclusively pursuing personal gain without considering coalition welfare or overall system performance, thereby demonstrating the advantages of FedCure's cooperative approach.

\subsubsection{Validation of Preference Rule Superiority}
\begin{figure}[tbp]
    \centering
    \begin{minipage}{0.48\textwidth}
        \centering
        \includegraphics[scale=0.11]{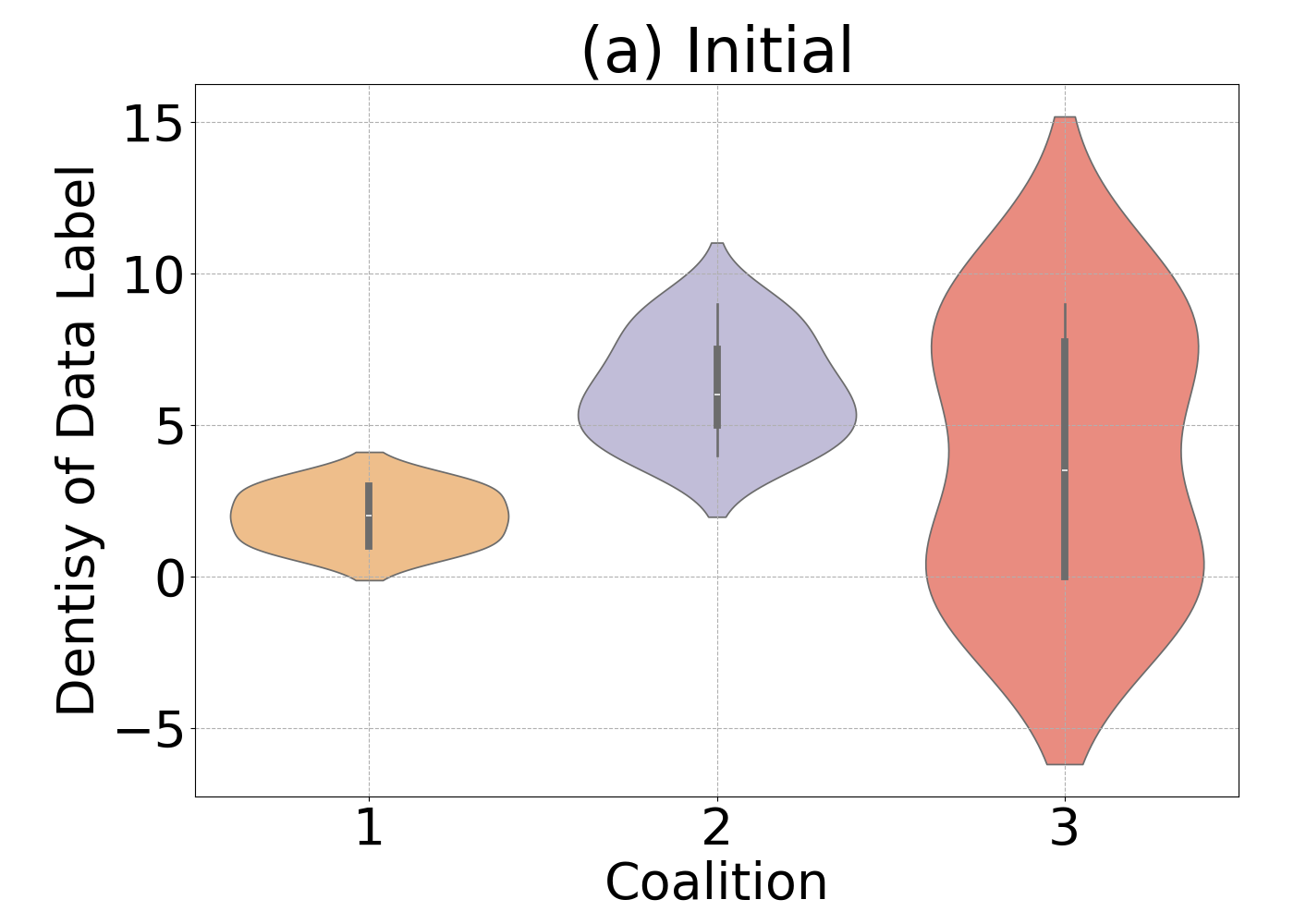}
        \includegraphics[scale=0.11]{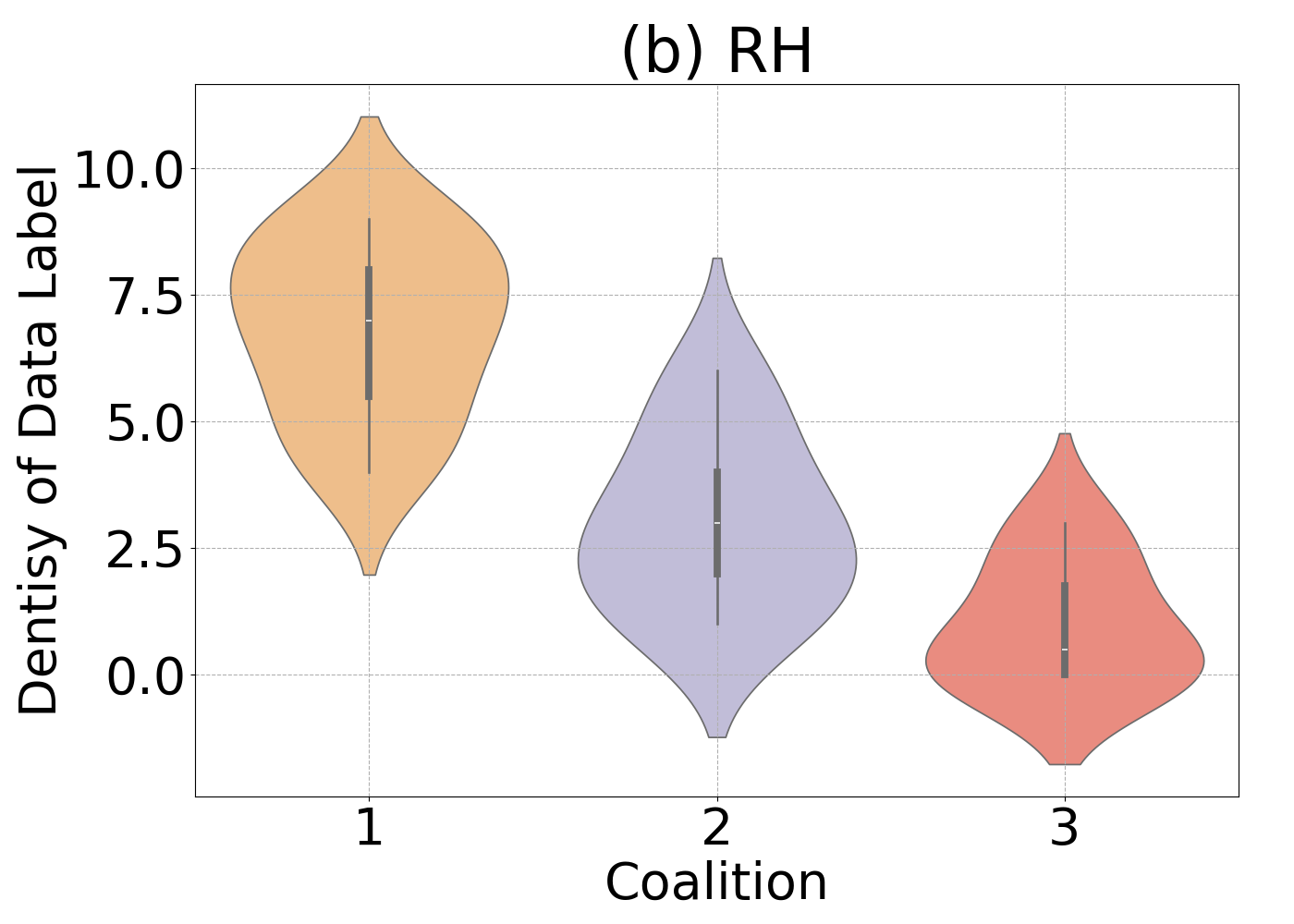}
    \end{minipage}
    
    \vspace{0.5em} 
    
    \begin{minipage}{0.48\textwidth}
        \centering
        \includegraphics[scale=0.11]{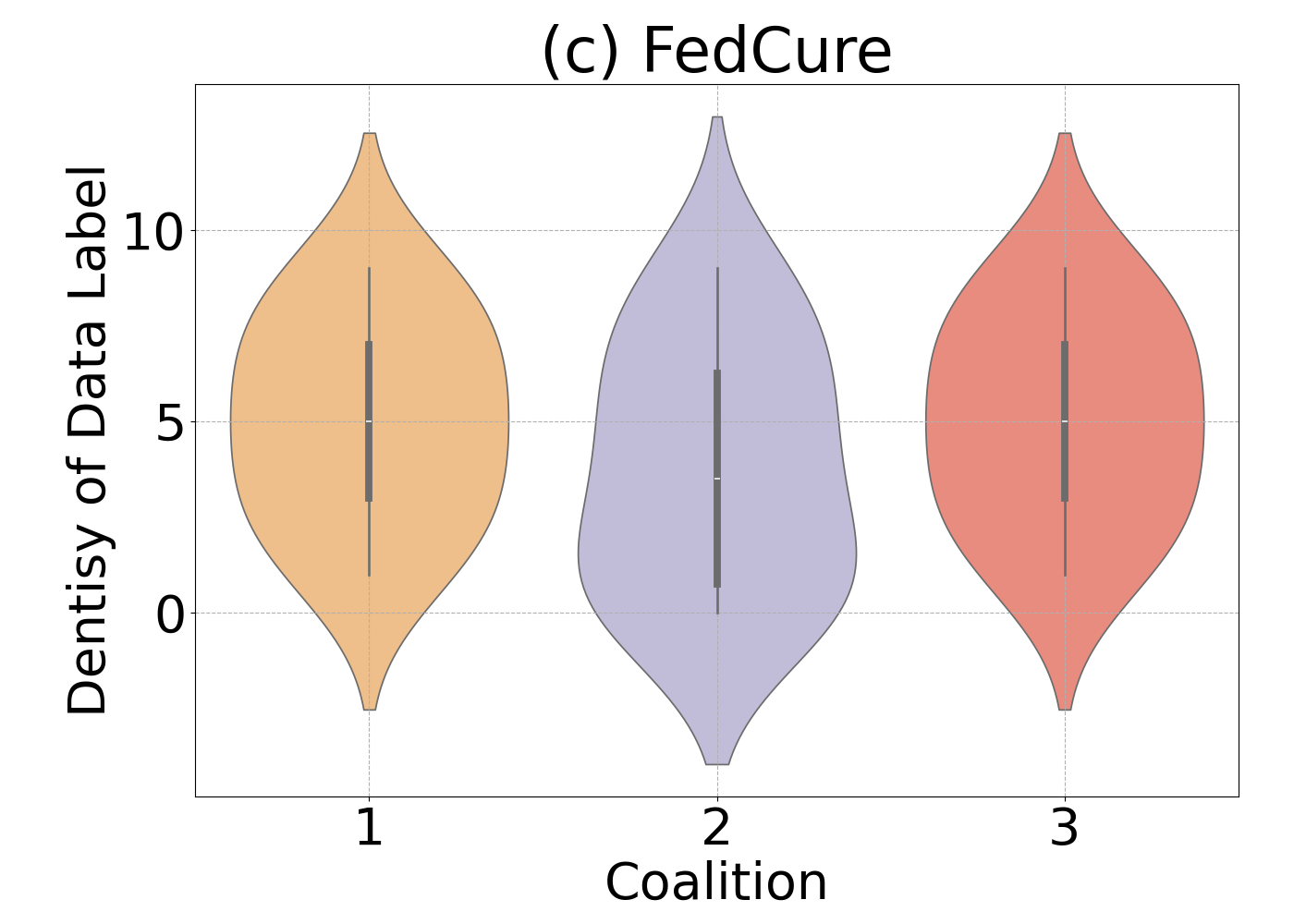}
        \includegraphics[scale=0.11]{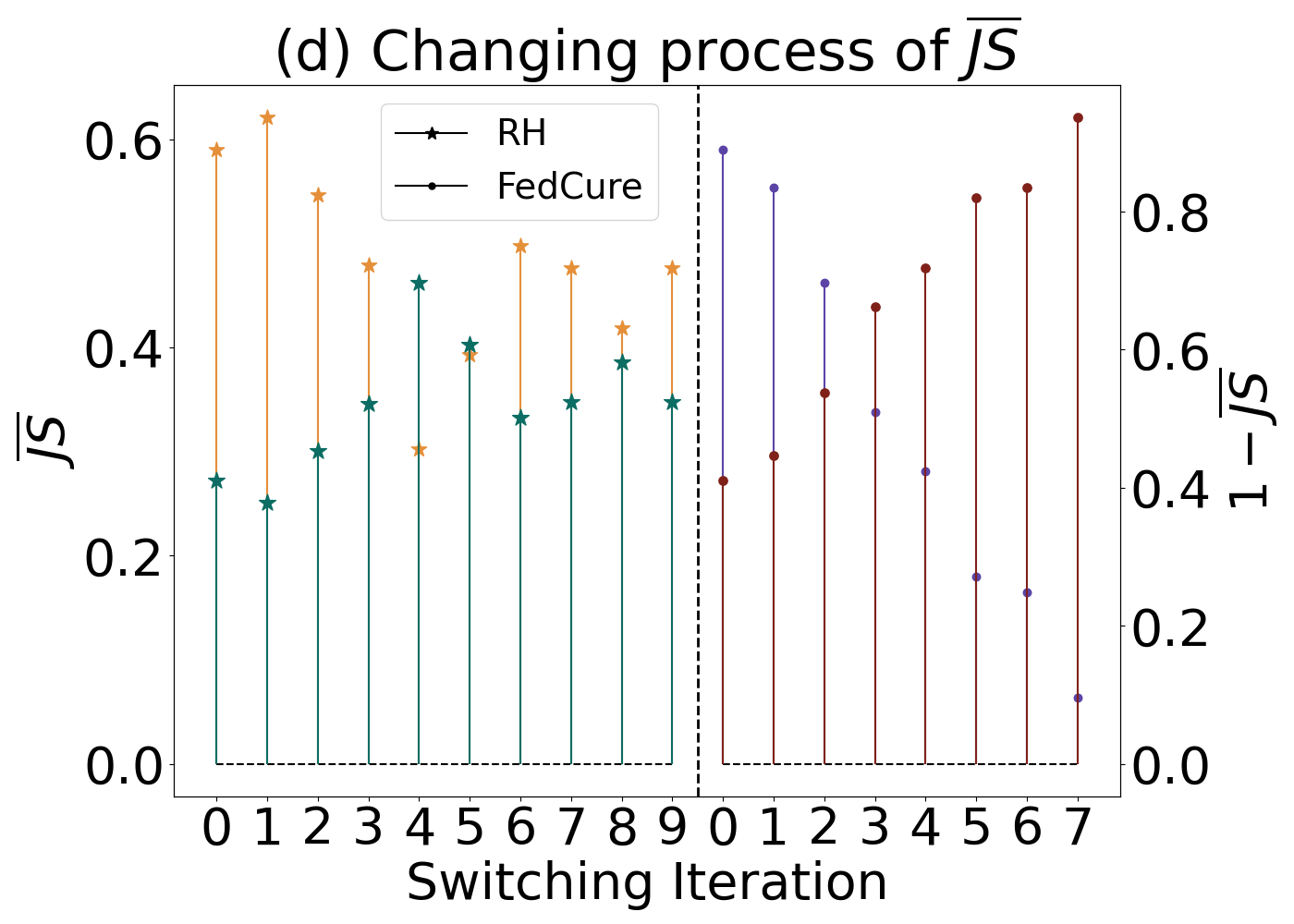}
    \end{minipage}
    
    \caption{Data Distribution adjustment by RH and FedCure.}
    \label{distribution}
\end{figure}

Figure \ref{distribution} presents the evolution of data distribution using the experimental framework (10 clients, 3 ESs) from (Ng et al. 2022), comparing RH and FedCure approaches. While both methods start with similar initial distributions (Fig. 5a), RH results in persistently dispersed data with distinct clustering patterns (Coalition 1:7.5, Coalition 2:2.5, Coalition 3:0 in Fig. 5b). In contrast, FedCure achieves superior balance (Fig. 5c) through its preference rule $\Upsilon_p$, which ensures: (1) monotonic $ \overline{\mathcal{JS}}$ reduction (right panel of Fig. 5d), unlike RH's unstable progression (left panel); (2) guaranteed improvement with each client switch; and (3) faster convergence (7 iterations total). These results demonstrate FedCure's effectiveness in optimizing data distribution through systematic coalition formation.

\subsubsection{Validation on real-world datasets} 
Figure 6 demonstrates FedCure's superior performance across four benchmark datasets (MNIST, CIFAR-10, SVHN, and CINIC-10), achieving significant accuracy improvements of 4.34\%, 8.20\%, 6.69\%, and 7.19\% respectively compared to the RH baseline. These consistent gains across diverse datasets validate FedCure's effectiveness in optimizing data distribution for enhanced model performance. The performance disparity stems from fundamental differences in coalition formation strategies: while RH's selfish client preference rules focus solely on individual utility maximization, neglecting system-wide impacts on data distribution, FedCure's cooperative approach actively optimizes global data distribution across ESs through intelligent coalition formation. This systematic optimization of data partitioning directly contributes to FedCure's superior model accuracy.

\begin{figure}[tbp]
    \centering
    \begin{minipage}{0.48\textwidth}
        \centering
        \includegraphics[scale=0.11]{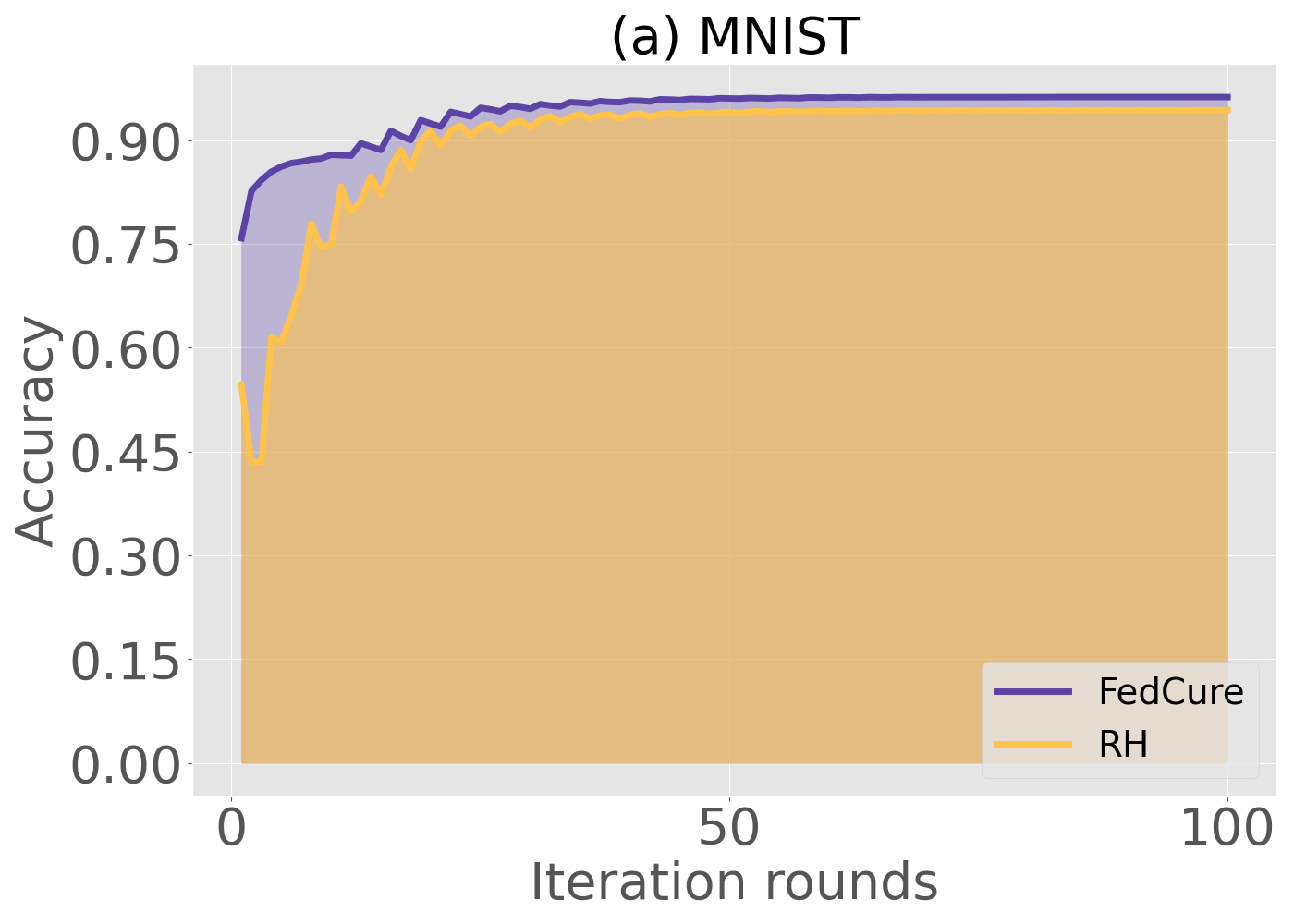}
        \includegraphics[scale=0.11]{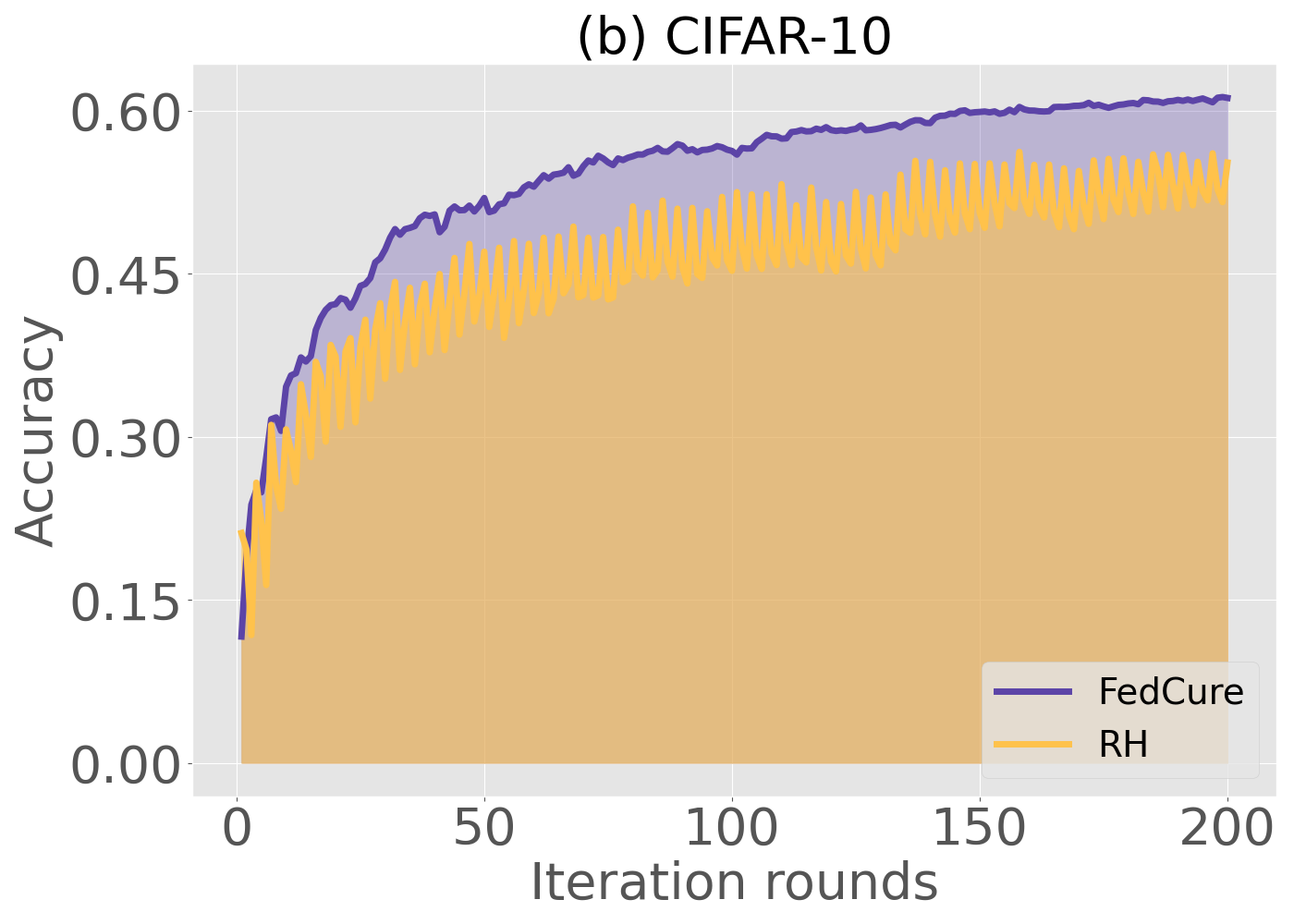}
    \end{minipage}
    
    \vspace{0.5em} 
    
    \begin{minipage}{0.48\textwidth}
        \centering
        \includegraphics[scale=0.11]{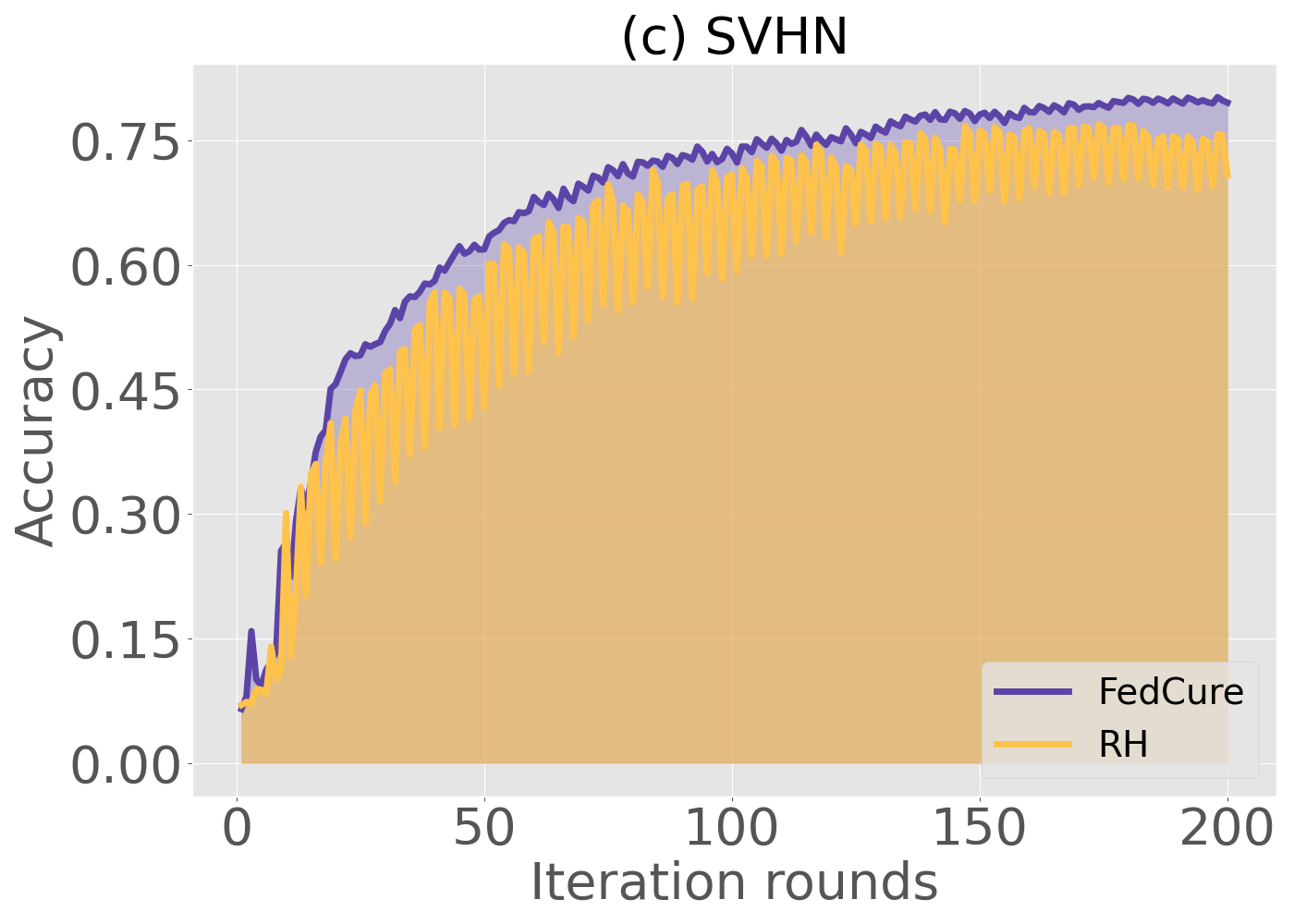}
        \includegraphics[scale=0.11]{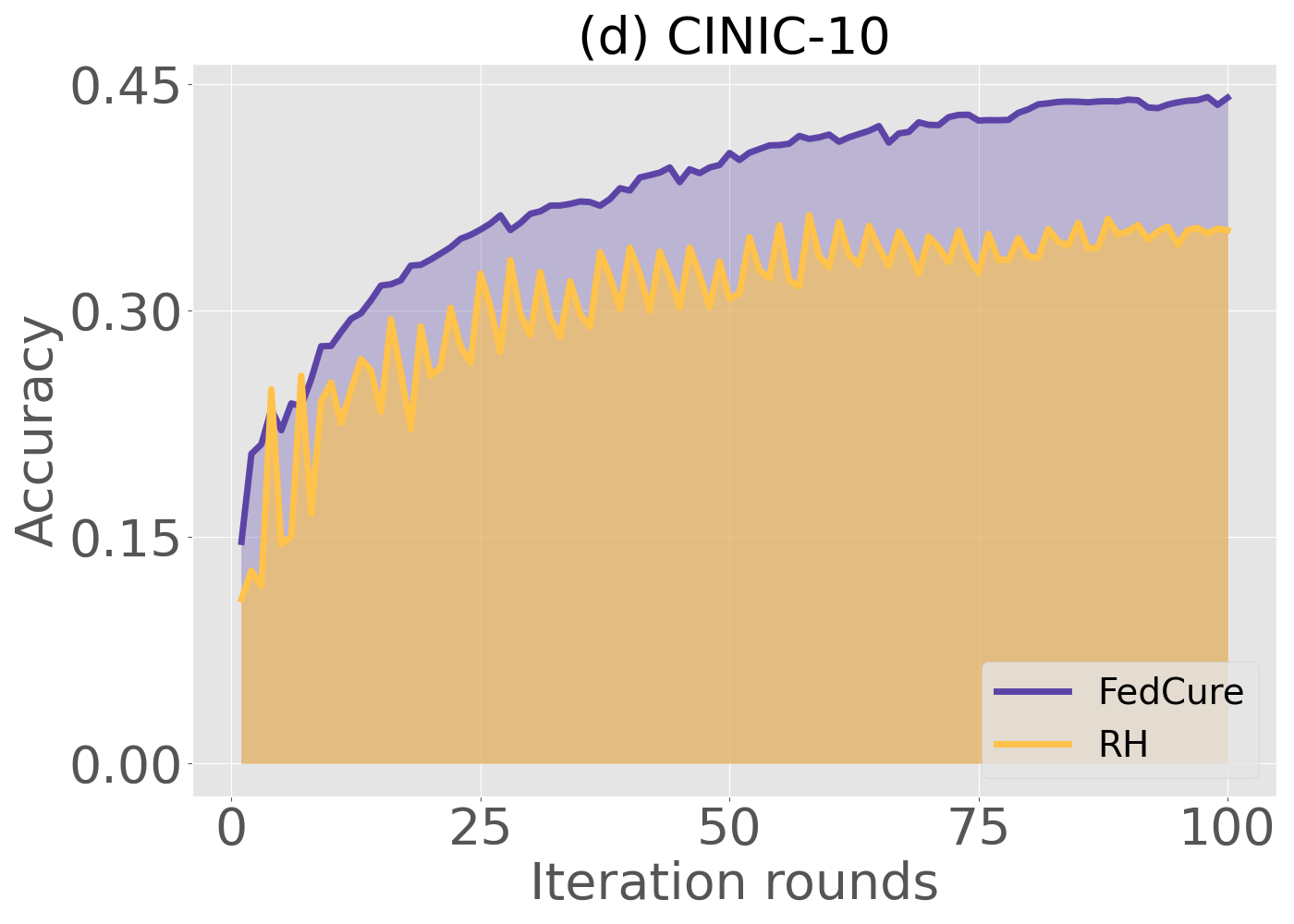}
    \end{minipage}
    
    \caption{Accuracy comparison among four datasets.}
    \label{acc}
\end{figure}

\bigskip

\bibliography{aaai2026}

\end{document}